\pdfoutput=1

\documentclass[11pt]{article}

\usepackage{acl}
\usepackage{graphicx}
\usepackage{tabularx}
\usepackage{amsfonts}
\usepackage{fontawesome5}
\usepackage{times}
\usepackage{latexsym}
\usepackage{amsmath} 
\usepackage{amsthm}

\usepackage{subcaption} 
\newcommand{\fa}[1]{\textcolor{green}{#1}}

\usepackage[T1]{fontenc}

\usepackage[utf8]{inputenc}
\usepackage{enumitem}
\setlist[itemize]{noitemsep, nolistsep}

\newcommand{\miniskip}{\vspace*{-.9\baselineskip}}
\newcommand{\shrink}{\vspace*{-.9\baselineskip}}

\usepackage{microtype}

\usepackage{inconsolata}

\title{SPILL: Domain-Adaptive Intent Clustering based on Selection and Pooling with Large Language Models 
} 

\author{I-Fan Lin \\
  Leiden University \\
  Leiden,  The Netherlands \\
  \texttt{\fontsize{9pt}{10pt}\selectfont i.lin@liacs.leidenuniv.nl} \\\And
  Faegheh Hasibi \\
  Radboud University \\
  Nijmegen, The Netherlands\\
  \texttt{\fontsize{9pt}{10pt}\selectfont faegheh.hasibi@ru.nl} \\\\\And
  Suzan Verberne \\
  Leiden University \\
  Leiden,  The Netherlands\\
\texttt{\fontsize{9pt}{10pt}\selectfont s.verberne@liacs.leidenuniv.nl} \\}

\begin{document}
\maketitle
\begin{abstract}

In this paper, we propose Selection and Pooling with Large Language Models (SPILL), an intuitive and domain-adaptive method for intent clustering without fine-tuning.
Existing embeddings-based clustering methods rely on a few labeled examples or unsupervised fine-tuning to optimize results for each new dataset, which makes them less generalizable to multiple datasets. Our goal is to make these existing embedders more generalizable to new domain datasets without further fine-tuning.
Inspired by our theoretical derivation and simulation results on the effectiveness of sampling and pooling techniques, we view the clustering task as a small-scale selection problem. A good solution to this problem is associated with better clustering performance. Accordingly, we propose a two-stage approach: First, for each utterance (referred to as the seed), we derive its embedding using an existing embedder. Then, we apply a distance metric to select a pool of candidates close to the seed. Because the embedder is not optimized for new datasets, in the second stage, we use an LLM to further select utterances from these candidates that share the same intent as the seed. Finally, we pool these selected candidates with the seed to derive a refined embedding for the seed.
We found that our method generally outperforms directly using an embedder, and it achieves comparable results to other state-of-the-art studies, even those that use much larger models and require fine-tuning, showing its strength and efficiency.
Our results indicate that our method enables existing embedders to be further improved without additional fine-tuning, making them more adaptable to new domain datasets. Additionally, viewing the clustering task as a small-scale selection problem gives the potential of using LLMs to customize clustering tasks according to the user's goals.\footnote{The source code is available: \url{ https://github.com/tom192180/SPILL_Clustering_LLM}}
\end{abstract}

\section{Introduction}
Intent detection is a fundamental component in task-oriented dialogue (TOD) systems, aimed at classifying user utterances into pre-defined intent categories \cite{ni2023recent}. Although some research has focused on addressing data scarcity \cite{siddique2021generalized, lin2024generate}, pre-defined intent labels are insufficient for addressing all user requests, as new intents emerge with growing complexity of customer requirements and appearance of novel domains on the business front. 
While the progress of transformer-based models has greatly enhanced intent detection performance, the identification of emerging intents in task-based conversational agents continues to present a challenge \cite{zhou2023probabilistic, rodriguez2024intentgpt}.

To address this issue, the majority of research aims to develop embedding models that group unlabeled utterances into clusters based on a labeled or unlabeled in-domain dataset \citep{zhang2021discovering, mou2022watch, zhang2023clustering, liang2023clusterprompt}. The goal of these approaches is to enable the embedder to learn a robust representation of user utterances while aligning with the cluster objective. Contrastive learning is commonly employed for this purpose, aiming to learn a representation through comparison 
\citep{le2020contrastive}. The clustering objective is achieved through the design of a cluster loss function \citep{zhang2021supporting, mou2022disentangled,du2023two}. 
Although these approaches yield good results, they require fine-tuning for each dataset. \citep{zhang2021supporting}. 

In recent years, advancements in generative Large Language Models (LLMs) \citep{touvron2023llama, riviere2024gemma} have facilitated improvements in intent clustering. \citet{zhang2023clusterllm} and \citet{liang2024synergizing} used LLMs to guide the fine-tuning of embedders, aiming to align the embeddings' clustering outcomes with LLMs' interpretations. 
Although these studies achieve state-of-the-art results, they face two primary challenges: First, modifying the loss function adds complexity, as it involves designing different loss functions and tuning additional hyperparameters, such as the weight of each loss term. This makes optimization more difficult. Second, building a new embedder requires optimization for each dataset, which limits generalizability. 

In this paper, we propose a theoretical framework for clustering, grounded by formal proofs, and an intuitive and effective approach to address these challenges. Our approach has three key goals: easy implementation, no need for fine-tuning, and the ability to adapt to unseen datasets. Our approach stays close to a theoretical rationale and we confirm its potential by simulation analysis. The key idea is that if we can identify a few utterances that share the same cluster as the seed utterance from a randomly selected subset, pooling the seed utterance with these selected utterances will bring them closer to the cluster centroid. Based on this premise, we can see a clustering task as a small-scale selection problems.

Our approach consists of two stages: In the first stage, for each utterance (referred to as the "seed utterance"), we use an existing embedder (a traditional encoder or a decoder-only LLM) to gain a larger pool of similar utterances. In the second stage, we use LLMs to further select utterances that share the same intent cluster as the seed utterance.

Note that our proposed approach is not intended to compete with other embedders. Instead, it is designed to complement most existing approaches and can strengthen each of the existing models. With our experiments, we show that our method can boost performance on the clustering task irrespective of the used embedder. In summary, we make three contributions: (1) we provide a theoretical framework supported by formal proofs and simulations, which frame the clustering task as a small-scale selection problem, providing both theoretical and empirical contributions to the task. (2) We propose a novel and easy-to-implement approach  that is generalizable, regardless of the embedders used and does not require fine-tuning and can operate with low computational resources; (3) We show our method enables domain adaptation in clustering for unseen datasets, achieving state-of-the-art results on four benchmark collections.

\section{Related Work}

\paragraph{Intent clustering with contrastive learning.}
Grouping user utterances and identifying new intents is essential in TOD systems \citep{soudani2024survey, zhang2021supporting, zhang2021discovering, mou2022watch, liang2023clusterprompt,du2023two}. Most research has focused on developing embedding models to create strong representations of user utterances. For instance, \citet{zhang2021discovering} pretrained a model with little labelled data and use k-means to produce cluster assignments as pseudo labels. They learn the intent representations under the supervision of the aligned pseudo-labels. 
\citet{zhang2021supporting} propose a method that optimizes both the contrastive loss and clustering loss together to build a sentence embedding model. To prevent overfitting on in-domain data during contrastive loss optimization, \citet{mou2022watch} limit the comparison to k-nearest neighbors instead of considering all possible neighbors. 
As earlier research focused on contrastive learning without fully accounting for the semantic meanings of labels, \citet{liang2023clusterprompt} use two-level contrastive learning to learn representations. This approach first aligns embeddings with several contrastive objectives, including their proposed label semantic alignment, then applies soft prompting to enhance the use of semantic knowledge in intent discrimination.


\paragraph{Sentence embeddings with LLM feedback.}
With the advancement of LLMs, research is increasingly using their capabilities to improve embedding.  \citet{zhang2023clusterllm} introduce a method that constructs multiple triplet questions, each consisting of an anchor data point and two candidate points. The triplets are initially selected from different clusters using a smaller embedder, and the LLM is then tasked with identifying the positive pair for the anchor point. After fine-tuning the embedder, they perform an initial clustering using the updated embeddings and then leverage the LLM to refine the clustering granularity. Their results show substantially better performance than traditional embedding methods like SCCL~\citep{zhang2021supporting} while using less data for training.  \citet{liang2024synergizing} leverage LLMs to derive intent descriptors. They then design contrastive loss functions to optimize a smaller embedder, synergizing LLMs and smaller language models for intent recognition. Instead of fine-tuning an embedder, \citet{viswanathan-etal-2024-large} use LLMs to improve the utterances by having them generate key phrases for each sentence, which are then added to the sentence and encoded into embeddings.  \citet{de2023idas} select prototypical utterances, generate labels for non-prototypical ones using LLMs, and encode both utterances and labels together.

\paragraph{Sentence embeddings in LLMs.}

While previous research has focused on methods that rely on pretrained contrastive loss embedders with feedback from LLMs, recent studies have shown that directly extracting embeddings from LLMs can also be effective. \citet{jiang2024scaling} propose an in-context learning approach to improve embeddings by introducing a ``one-word limitation.'' The idea is to instruct LLMs to summarize the input sentence into a single word. They found that this approach can still achieve good performance without fine-tuning. \citet{springer2025repetition} propose the echo embedding approach. Because auto-regressive embeddings do not capture context from later tokens, they pass the sentence through the model twice and pool embeddings only from the second occurrence. Their experiment shows that this method outperforms traditional pooling. \citet{lei2024meta} propose a meta-task prompting method with a one-word limitation. The embeddings are created using a series of carefully designed prompts that cover different aspects of meaning, using the one-word limitation to improve the embeddings.

\section{Theoretical Framework}
In this section, we introduce our theoretical derivation, followed by validations through empirical simulations.
\subsection{Problem formulation}

We cast the problem of identifying emerging intents to a clustering task, where conversation utterances are grouped into clusters, with each cluster corresponding to a newly identified intent.
Consider a collection of data points $D = \{x_{i}\}_{i=1}^{N}$, where each data point $x_{i} \in D$ corresponds to an utterance, and $N$ is the total number of data points. The task is to partition $D$ into cluster sets  $\{\hat{S}_{l}\}_{l=1}^{\hat{M}}$, where $\hat{M}$ is the number of the unique clusters. Note that $\sum_{l=1}^{\hat{M}} |\hat{S}_{l}| = N$. The objective is to ensure that the clustering results $\{\hat{S}_{l}\}_{l=1}^{\hat{M}}$ are as close as possible to the true partition $\{S_{l}\}_{l=1}^{M}$, where $M$ is the number of unique clusters in the ground truth. Similarly, $\sum_{l=1}^{M} |S_{l}| = N$.

\subsection{Theoretical grounding and proof}

For simplicity in the theoretical development, we assume sampling is performed with replacement.
Consider cluster $S \in \{S_{l}\}_{l=1}^{M}$, containing $N_S$ data points such that $S = \{x_{i}\}_{i=1}^{N_{S}}$.
For each data point $x_{i}$, we derive a $d$ dimensional vector representation $\mathbf{z}_{i} \in \mathbb{R}^d$ from an embedder. This results in the set $\{\mathbf{z}_{i}\}_{i=1}^{N_{S}}$. We denote $\sigma_h^{2}$ and $\mu_h$ as the variance and mean of the cluster for the $h$-th dimension of our $d$-dimensional space. $\mu_h$ is also the cluster centroid in $h$-th dimension. If we randomly select one utterance from the cluster $S$, represented by its corresponding vector 
${\mathbf{Z}}_{i}$,\footnote{We use capital to highlight it is a random vector.} then we randomly select $k$ elements $\mathbf{Z}_{i1}, \mathbf{Z}_{i2}, ..., \mathbf{Z}_{ik} $ with replacement  from $\{\mathbf{z}_{i}\}_{i=1}^{N_{S}}$ to derive its pooling version $\overset{*}{\mathbf{Z}}_{i}$, defined as:

\begin{align}\label{pooling}
 \overset{*}{\mathbf{Z}}_{i} :=  \frac{\mathbf{Z}_{i} + \sum_{m = 1}^{k} \mathbf{Z}_{im}}{1 + k}.
\end{align}

Based on this, we formally prove that the following inequality holds:

\begin{align}\label{inequality}
\hspace{-0.5cm}
\sum_{h=1}^{d}E[(\overset{*}{Z}_{ih} - \mu_h)^2] - E[(Z_{ih} -\mu_h)^2] < 0,
\end{align}
where $Z_{ih} \in R$ and $\overset{*}{Z}_{ih} \in R$ represent elements of the $h$-th dimension of $\mathbf{Z}_{i}$ and $\overset{*}{\mathbf{Z}}_{i}$, respectively. 
This inequality suggests that using this sampling and pooling approach, the samples are closer to the cluster centroid, which implies that clustering can be approached as a small-scale selection task. 

\begin{proof}

To show the inequality holds, we first to show both $\overset{*}{Z}_{ih}$ and $Z_{ih}$ have the same mean:

\begin{align*}
 E(\overset{*}{Z}_{ih}) &=  E\bigg( \frac{{Z}_{ih} + \sum_{m = 1}^{k} {Z}_{imh}}{1 + k} \bigg)\\
 &= \frac{1}{(1+k)}\bigg\{E({Z}_{ih}) + \sum_{m = 1}^{k} E({Z}_{imh})\bigg\}
\\
&= \frac{1}{(1+k)} (\mu_{h} + k\mu_{h})\\
& = \mu_{h}
\end{align*}

${Z}_{imh} $ is the $h$th dimension element of $\mathbf{{Z}}_{im}$ (see Eq. \ref{pooling}). Since $ E[\overset{*}{Z}_{ih}] = E[Z_{ih}] = \mu_{h}$, the inequality (see the formula \ref{inequality}) can be rewritten as follows:

\begin{align}\label{var}
\sum_{h=1}^{d} \text{Var}[\overset{*}{Z}_{ih}] -\text{Var}[Z_{ih}] < 0
\end{align}
Then, we have:

\begin{align*}
\text{Var}[\overset{*}{Z}_{ih}]&= \text{Var}\bigg[\frac{Z_{ih} + \sum_{m = 1}^{k} Z_{imh}}{1 + k}\bigg]\\
&\overset{\diamond}= \frac{1}{(1+k)^2}\bigg[\text{Var}[Z_{ih}] + \sum_{m = 1}^{k} \text{Var}[Z_{imh}]\bigg]\\
&= \frac{1}{(1+k)^2}[\sigma_{h}^{2} + k \cdot \sigma_{h}^{2}]\\
&= \frac{1}{1+k}\cdot \sigma_{h}^{2} < \sigma^{2}_{h} = \text{Var}[Z_{ih}]
\end{align*}

\end{proof}
Note that the step marked with  $\diamond$ holds due to the independence of these random variables. Since this property holds for each dimension, the entire summation (see Eq. \ref{var}) will hold as well.

According to this inequality, instead of identifying all similar utterances at once, we only need to determine whether a randomly chosen subset of utterances belongs to the same cluster as the seed utterance. The idea based on the inequality is that if we can perform selection perfectly on the subset of data points, then any cluster will show reduced variance when each data point is pooled with randomly selected points from the same cluster, compared to the variation in the original points. We examine the relationship between variance and clustering performance in Section~\ref{sec:simulation} using simulations.

\subsection{Simulation study} \label{sec:simulation}

Inspired by the theoretical principles outlined above, our main goal is to identify utterances that share the same intent as the seed utterance. To achieve this, we only need to consider a randomly selected subset of candidate utterances at a time.

A straightforward way to compare the seed utterance with a subset is to use an LLM directly. However, choosing the right subset size is challenging: a small subset may miss related utterances, while a large one increases computational cost. To address this, we use an embedding model to select a subset of similar utterances as a starting point. However, this approach might deviate from the theory's assumption that the subset is randomly selected from the whole cluster, as it focuses only on locally close utterances. To evaluate the impact brought by embedder-base selection, we conduct a simulation study.

We consider 3 clusters in a 128-dimensional space. The mean and variances are averaged over 128 dimensions. For each experiment trial, between 50 and 250 data points are independently sampled for each cluster. The data is drawn from either a normal distribution or a skewed log-normal distribution. For the normal distribution, the $\mu$ and $\sigma^2$ in each dimension are sampled from the ranges $(0, 1e-10)$ and $(20, 60)$, respectively. The log-normal distribution is generated by taking the exponential of data points sampled from the normal distribution, with the $\sigma^2$ and $\mu$ in each dimension sampled from the ranges $(0, 1e-10)$ and $(1.5, 2)$.

In the simulation experiments, we analyze two extreme pooling strategies: (1) randomly selecting $k$ utterances from the same cluster to pool with the seed utterance (strategy \textbf{Rd}), following the theoretical analysis, and (2) selecting only the top-$k$ closest utterances from the same cluster to pool with the seed utterance (strategy \textbf{TopK}). Sampling is performed without replacement to prevent over-representation of certain examples, which can occur in small clusters. 

Table \ref{tab:simulation} shows that both \textbf{Rd} and \textbf{TopK} pooling strategies reduce variances while simultaneously improving clustering metrics in both non-skewed and skewed settings. Furthermore, the results indicate that increasing the pooling size leads to better clustering performance. Finally, even in scenarios with a skewed distribution, both pooling methods continue to enhance performance. We also investigate whether the results hold with varying dimensionality. We found that for higher dimensionality, the variances for both \textbf{Rd} and \textbf{TopK} remain reduced and the clustering performance improved (see Appendix \ref{simu_dim} for more details). 

The findings of our simulation study are as follows: (1) lower variance within a cluster is associated with better clustering performance, (2) increasing the value of $k$ tends to improve clustering performance, and (3) pooling based on the closest $k$ still shows improvement, indicating that using an embedder as an initial selection step is a reasonable approach.

\begin{table}[t]
  \centering
  \setlength{\tabcolsep}{2pt}
  \footnotesize
    \begin{tabular}{l|c|cc|cc|cc}
    \hline
      & k = 0 &\multicolumn{2}{c|}{k = 5} & \multicolumn{2}{c|}{k = 10}  & \multicolumn{2}{c}{k = 20} \\
     \cline{1-4} \cline{5-6} \cline{7-8}  
     &   & Rd & TopK   & Rd & TopK    & Rd & TopK  \\
    \hline
    \textbf{Normal} &&&&&&&\\
    \hspace{0.4cm}$\hat{var}$ &39.76 & 6.33 & 7.70 & 3.30 & 4.20 & 1.58 & 2.12 \\
    \hspace{0.4cm}Acc &37.54 & 69.73 & 65.74 & 94.71 & 92.61 & 99.54 & 99.20 \\
    \hspace{0.4cm}NMI &0.52 & 33.15 & 32.94 & 80.32 & 77.32 & 97.60 & 96.33 \\
    \hline
    \textbf{Log-normal} &&&&&&&\\
    \hspace{0.4cm}$\hat{var}$ &28.12 & 4.45 & 3.04 & 2.33 & 1.29 & 1.12 & 0.47 \\
    \hspace{0.4cm}Acc &43.52 & 47.54 & 53.50 & 58.60 & 88.16 & 81.67 & 98.77 \\
    \hspace{0.4cm}NMI &0.93 & 9.82 & 19.14 & 31.63 & 71.41 & 74.94 & 94.16 \\
    \hline
  \end{tabular}

  \caption{Average estimated variances, mean, and clustering metrics over 50 runs in our simulation study. $k$ is  number of data points selected to pool with the seed, and $\hat{var}$ denotes estimated variance for the one of the cluster over 128 dimensions; we report only one of them since all clusters have the same trend.}
  \miniskip
  \label{tab:simulation}
\end{table}

\section{Computational Method}

Based on our theoretical analysis and simulation results, randomly selecting utterances among the entire cluster to pool with a seed utterance is guaranteed to reduce clustering variance, which is associated with better clustering performance. The simulation results further demonstrate that pooling using top$-k$ selection can still outperform no pooling and can approach the performance of random selection achieved with larger $k$ in higher dimensional spaces. Building on these findings, we propose a two-stage selection approach. In the first stage, we implement a selection strategy based on these findings. In the second stage, to refine the candidates from the first stage, we leverage the capabilities of LLM to identify the best utterances in the selection, ensuring that the selected utterances belong to the same cluster.

\subsection{First stage: Embedding based selection}

In this stage, we use open-source models to extract the utterance embeddings: we feed $x_{i}$ into a existing pretrained encoder or an LLM (referred to as ``embedder'' in the following text) to extract its embedding, denoted as $\mathbf{z}_{i} := \text{Embedder}(x_{i})$. Let $d(\mathbf{z}_{i},\mathbf{z}_{j})$ be a distance function that compute the distance between the two embeddings $\mathbf{z}_{i}$ and $\mathbf{z}_{j}$, where $i \neq j$. We use Euclidean distance and denote the distance as $d_{ij}$. By sorting the distances, we can select the top $l_{top}$ closest utterances to the seed $x_{i}$, which have highest chance of being in the same cluster. These corresponding utterances are denoted as $\{x_{ij}\}_{j=1}^{l_{top}}$. Additionally, to introduce some variety, we also select utterances $l_{random}$ $\{x_{ij}\}_{j=1}^{l_{random}}$ using chunk sampling: we split the whole sample into $l_{random}$ chunks, and take the first closest utterance from each chunk. This chunk sampling method introduces variety while ensuring that relatively close utterances are selected. Thus, the entire candidate set is given by $D_{i} := \{x_{ij}\}_{j=1}^{l_{top}} \cup$ $\{x_{ij}\}_{j=1}^{l_{random}}$.

The goal of the first stage is to quickly select utterances that are more likely to belong to the same cluster as the seed $x_{i}$. However, since the embedder is not specifically optimized for the datasets we are experimenting with, and the clusters may be in overlapping areas in the vector space, it is possible that some candidates belong to a different cluster than the seed $x_{i}$.

\subsection{Second stage: LLM based selection }

To address the possibility that the candidate set $D_{i}$ may include utterances from different clusters than $x_{i}$, we use in-context learning with an LLM to select utterances that share the same intent as the seed utterance from $D_{i}$. Note that $D_{i}$ is shuffled before being fed into the LLM. These selected utterances are denoted as $\tilde{D}_{i}$, note $\tilde{D}_{i} \subseteq$ $D_{i}$. We then take the seed $x_{i}$ along with all elements from $\tilde{D}{i}$ to compute the mean pooling. This pooled representation, $\mathbf{\overset{*}z}_{i}$, will be used for the clustering algorithm. The size of $\tilde{D}_{i}$ varies for each seed utterance, as it depends on the LLM's selection result.

The purpose for the second stage is to leverage the power of LLMs with a simple designed prompt to further select the utterances sharing the same intent as the seed utterance.


\section{Experimental Settings}


\subsection{Datasets and models} 

We use \textbf{SGD} \cite{rastogi2020towards}, \textbf{Bank77} \citep{Casanueva2020}, \textbf{CLINC150} \cite{larson2019evaluation}, \textbf{Mtop} and \textbf{Massive} for our experiments. Statistics are provided in Table \ref{tab:dataset}. 

In the first stage, embeddings extraction, we experiment with three decoder models \textbf{Qwen2.5-7B-Instruct} \cite{qwen2.5}, \textbf{Llama3.1-8B-Instruct} \cite{llama3modelcard}, and \textbf{Gemma-2-9b-it} \cite{riviere2024gemma}, as well as two encoder models \textbf{E5-large} \cite{wang2022text} and \textbf{Instructor-large} \cite{su2023one}. In the second stage, we use an LLM to verify utterances. If a decoder model is used in the first stage as an embedder, we continue with the same model in the second stage; for the encoder models in the first stage, we tried all the three LLMs in the second stage.  For simplicity, we refer to these models with shortened names: Qwen, Llama, Gemma,  E5, and Instructor in the following sections. 

\begin{table}[t]
\centering
\small
\begin{tabular}{c|c|c}
\hline
Dataset & \# clusters & \# utterances \\ 
\hline
\textbf{SGD}  & 46   & 57.2K$^{*}$           \\ 
\textbf{Bank77}     & 77     & 3,080     \\ 
\textbf{Clinc150}    &150     & 4,500      \\ 
\textbf{Mtop}& 102     & 4,386     \\ \
\textbf{Massive}   & 59     & 2,974     \\ 
\hline
\end{tabular}
\caption{Dataset Statistics. Note that for \textbf{SGD}, we use only a partial set (9.6K) for setting the hyperparameters. For other datasets, we use the same settings as \citet{zhang2023clusterllm}.}
\label{tab:dataset}
\end{table}

\subsection{Prompts}
\paragraph{First stage.} We build on previous research to implement our method for extracting embeddings. The prompts we used can be seen in Table \ref{tab:1stpt}. We follow \citep{jiang2024scaling, springer2025repetition}, and apply minor modifications to include wording tailored to our specific task of intent detection. For \textbf{Instructor}, we adopt the same prompt used by \citet{zhang2023clusterllm},  which is similar to the ones used in \citep{su2023one}. For \textbf{E5-large}, we follow \citep{wang2022text} and add `Query:' as prefix. 

\begin{table}[t]
\centering
\small
\begin{tabular}{|p{7cm}|}
\hline

\textbf{Summarizer} \scriptsize  \citep{jiang2024scaling}:\\

\scriptsize
The task is intent detection. The goal is to identify the purpose or goal behind a user input. The user intent of this sentence: \{sentences\} means in one word:"\\

\hline

\textbf{Echo (Repetition)} 
\scriptsize \citep{springer2025repetition}:\\
\scriptsize Instruct: The task is intent detection. The goal is to identify the purpose or goal behind a user input. Give the user intent of the utterance,\\

\scriptsize User utterances:\{sentences\} User utterances again:\{sentences\}\\

\hline

\textbf{Instructor} \scriptsize  \citep{su2023one} and \cite{liang2023clusterprompt}:\\

\scriptsize Bank77: Represent the bank purpose for retrieval:\\

\scriptsize Mtop: Represent the sentence for retrieval:\\

\scriptsize Clinc150 and Massive: Represent the sentence for retrieving the purpose:\\

\hline
\end{tabular}
\caption{Task Instructions Prompt for the first stage. Note that for \textbf{Summarizer} and \textbf{Echo}, we use the same prompt across different datasets.}
\label{tab:1stpt}
\end{table}

\paragraph{Second stage.} \label{2nd_stage_pt}
In this stage we use the same prompt across different LLMs, shown in Table \ref{tab:pt}.

\begin{table}[t]
\centering
\scriptsize
\shrink
\begin{tabular}{|p{7cm}|}
\hline
\\
\textbf{Task Instructions:} \\
\\
\textbf{Step 1: Identify Intent Clusters}\\
Review the Candidate Utterances to identify their individual intents and group them into clusters based on shared intent. Candidates may either align with the same cluster as the Target Utterance or belong to entirely different clusters.\\
Note: Intent refers to the request or the purpose the user wants to achieve.\\
\\
\textbf{Step 2: Match Intent with Target Utterance}\\ 
Compare each Candidate's intent to the Target Utterance, using the clusters you identified. Select only Candidates from the same intent cluster as the Target Utterance.\\
Note: Choose a Candidate only if its intent clearly aligns with the Target Utterance's purpose.\\
\\
\textbf{Answer Format:}\\
Only provide the final selection of Candidate Utterances by listing their numbers if they match the Target Utterance intent or request.
\begin{enumerate}[left=0pt, itemsep=0pt, topsep=0pt]
    \item If Candidates 3, 4, 9, and 11 match, write: The Candidate utterances numbers are: 3, 4, 9, 11
    \item If no Candidate matches, write: The Candidate utterances numbers are: none
\end{enumerate}
Note: Stick to the answer format and avoid providing extra explanations.\\
\\
\textbf{Task:}
\\
Target Utterance: \{sentence 1\} \\
Candidate Utterances:  \\
1. \{sentence 1\}\\
... \\
L. \{sentence L\}\\
\hline

\end{tabular}
\caption{Task Instructions Prompt. The boldface used here is for readability and is not used in the prompt.}
\label{tab:pt}
\shrink
\end{table}

\subsection{Embedding derivation} 

\textbf{E5} and \textbf{Instructor} use mean pooling to derive embeddings for sentence representation. For \textbf{Echo}, the paper shows that mean pooling over the final hidden layer achieves better performance. \textbf{Summarizer} uses the last token embedding from the final hidden layer. We adopt their settings for our follow-up experiments.

\subsection{Evaluation} 

To evaluate the embedding quality, we assume (for simplicity) that the true cluster number is known. This assumption aligns with with previous research evaluation practices \citep{zhang2021supporting, zhang2022new, zhang2023clusterllm, viswanathan-etal-2024-large, liang2024synergizing}, although our method does not require it.
Therefore, the number of clusters will correspond to the number of labels for each dataset, and the KMeans algorithm \citep{1056489} is applied for clustering.

Since the datasets used in our experiments are labeled, we apply standard clustering metrics to evaluate the results. These metrics include normalized mutual information (NMI) and clustering accuracy (Acc) \citep{rand1971objective, meilua2007comparing, huang2014deep, gung2023intent}. 

\subsection{Hyperparameters setting}


Although our proposed method does not require updating the model parameters, it relies on one key hyperparameter: the values used to select first-stage candidates, specifically the values of $l_{top}$ and $l_{random}$. We use \textbf{SGD} as an external dataset to optimize these parameters. This avoids directly applying our method to the evaluation datasets, ensuring a fair comparison. We set our \textit{tolerance} \footnote{we call tolerance because increasing it adds more computation in the second stage.}  to $ l_{top} + l_{random}=20$ and experiment with different combinations of $l_{top} \in \{2, 4, ..., 20\}$. We found that the combination $l_{top}=14; l_{random} = 6$ gives the best performance, though overall results show little difference across combinations; see more detail in the Appendix \ref{hyperparameter}. We fix the setting (14,6) throughout our experiments across all different datasets and models.

\subsection{Baselines} 
We compare our method with the state-of-the-art approaches, all of which operate under the same setting. We perform clustering directly on an unlabeled dataset, which itself serves as the test data without the need for any additional training dataset. Among these, \textbf{SCCL} and \textbf{ClusterLLM} involve fine-tuning of their embedding models.


\paragraph{KeyphraseCluster} \cite{viswanathan-etal-2024-large} uses \textbf{GPT} (gpt-3.5-turbo) to generate key phrases for each utterance, which are then concatenated for clustering. They use \textbf{Instructor-large} as their backbone of the embedder \citep{su2023one}.
 
\paragraph{ClusterLLM} \cite{zhang2023clusterllm} uses \textbf{GPT} (gpt-3.5-turbo) to construct hard triplets from the test dataset and fine-tune a small embedder. They use \textbf{Instructor-large} \citep{su2023one} and \textbf{E5-large} embedders \cite{wang2022text}. We also report their reproduced results for \textbf{SCCL} \citep{zhang2021supporting}, a classic approach in text clustering.

\paragraph{Echo and Summarizer} embeddings \cite{springer2025repetition, lei2024meta} do not rely on contrastive learning for optimization. Instead, they use their own prompts to derive the desired embedding from decoder-only LLMs.

\section{Results}

\begin{table*}[t]
  \centering
  \setlength{\tabcolsep}{2.5pt}
  \footnotesize
  \shrink
    \begin{tabular}{lcc|cc|cc|cc}
    \hline
    &\multicolumn{2}{c|}{Bank 77} & \multicolumn{2}{c|}{Clinc150}  & \multicolumn{2}{c|}{Mtop} & \multicolumn{2}{c}{Massive}\\
     \cline{1-3} \cline{4-5} \cline{6-7} \cline{8-9}  
    & NMI & Acc & NMI & Acc & NMI & Acc& NMI & Acc\\
    \hline
\textbf{Fine-tuned embeddings}&&&&&&&\\


    \hspace{0.4cm}SCCL$_{\text{E5}}$ & 77.34 \tiny{(0.62)}  &63.60 \tiny{(1.37)}&  91.89\tiny{(0.49)} & 77.96\tiny{(1.78)}&70.42\tiny{((0.34)}& 33.82 \tiny{((1.07)} &    71.57 \tiny{((0.89)} & 54.48\tiny{((1.80)}\\  

    \hspace{0.4cm}SCCL$_{\text{Instructor}}$ & 81.77 \tiny{(1.36)}  &65.48 \tiny{(1.36)}&  92.94\tiny{(0.44)} & 80.85\tiny{(0.74)}&73.52\tiny{((0.38)}& 34.28 \tiny{((0.58)} &    73.90 \tiny{((0.36)} & 54.10\tiny{((1.05)}\\  

        \hspace{0.4cm}ClusterLLM$_{\text{E5}}$
        (GPT)&  84.16 
        \tiny{(0.36)} &70.13 \tiny{(1.34)}    &  92.92 \tiny{(0.29)} &80.48 \tiny{(0.93)} & 74.46 \tiny{(0.11)} &37.22 \tiny{(1.18)}  &  74.39 \tiny{(0.21)} & 56.08 \tiny{(1.01)}\\

        \hspace{0.4cm}ClusterLLM$_{\text{Instructor}}$ 
        (GPT)&  \textbf{85.15} \tiny{(0.41)} &71.20 \tiny{(1.59)}    &  94.00 \tiny{(0.21)} &83.80 \tiny{(0.41)} & 73.83 \tiny{(0.79)} &35.04 \tiny{(0.97)}  &  77.64 \tiny{(0.21)} & 60.69 \tiny{(0.96)}\\
    \hline
    \textbf{E5-large}&&&&&&&\\
    \hspace{0.2cm}Plain  & 77.19 \tiny{(0.34)} & 59.60 \tiny{(1.42)} & 91.27 \tiny{(0.38)} & 75.92 \tiny{(0.91)} & 70.87 \tiny{(0.23)} & 34.21 \tiny{(0.57)} & 71.38 \tiny{(0.55)} & 53.85 \tiny{(1.28)} \\

    \hspace{0.2cm}SPILL (Gemma)& 83.56 \tiny{(0.47)} & 70.25 \tiny{(1.56)} & \underline{92.93} \tiny{(0.08)} & \underline{83.18} \tiny{(0.78)} & \underline{71.77} \tiny{(0.35)} & \underline{36.83} \tiny{(1.10)} & \underline{75.40} \tiny{(0.51)} & \underline{60.28} \tiny{(1.81)} \\
    \hspace{0.2cm}SPILL (LLama) & 80.94$^{}$ \tiny{(0.29)} & 66.84$^{}$ \tiny{(0.97)} & 91.41 \tiny{(0.07)} & 80.09$^{}$ \tiny{(1.31)} & 70.52 \tiny{(0.51)} & 35.04 \tiny{(1.06)} & 72.74$^{}$ \tiny{(0.44)} & 58.06$^{}$ \tiny{(1.52)} \\
    \hspace{0.2cm}SPILL (Qwen) & \underline{83.64} \tiny{(0.26)} & \underline{70.31} \tiny{(0.75)} & 92.67 \tiny{(0.31)} & 82.58 \tiny{(0.87)} & 71.20 \tiny{(0.36)} & 36.48 \tiny{(0.73)} & 74.46 \tiny{(0.57)} & 59.38 \tiny{(1.20)} \\

    \hline

    \textbf{Instructor-large}&&&&&&&\\
    \hspace{0.2cm}Plain  & 82.38 \tiny{(0.59)} & 65.70 \tiny{(1.78)} & 93.25 \tiny{(0.32)} & 81.12 \tiny{(2.27)} & 71.69 \tiny{(0.60)} & 34.06 \tiny{(2.51)} & 74.56 \tiny{(0.37)} & 56.62 \tiny{(1.75)} \\
    \hspace{0.2cm}KeyphraseClust. (GPT) & 82.4\phantom{0}  \tiny{(0.0)} &65.3\phantom{0}  \tiny{(0.0)}&  92.6\phantom{0}  \tiny{(0.0)}&79.4\phantom{0}  \tiny{(0.0)}&  - &-  &    - & -\\

    \hspace{0.2cm}SPILL (Gemma) & 85.01 \tiny{(0.29)} & 71.05 \tiny{(0.83)} & \underline{93.77}  \tiny{(0.35)} & \underline{85.14}  \tiny{(1.04)} & \underline{72.65}  \tiny{(0.32)} & \underline{37.11} \tiny{(1.62)} & \underline{77.62} \tiny{(0.46)} & \underline{62.42} \tiny{(2.06)} \\
    \hspace{0.2cm}SPILL (LLama) & 83.37 \tiny{(0.22)} & 69.55$^{}$ \tiny{(0.60)} & 92.96 \tiny{(0.18)} & 84.31 \tiny{(0.72)} & 71.41 \tiny{(0.31)} & 35.18 \tiny{(1.12)} & 75.28 \tiny{(0.83)} & 58.79 \tiny{(2.75)} \\
    \hspace{0.2cm}SPILL (Qwen) & \underline{85.12} \tiny{(0.30)} & \textbf{\underline{71.48}} \tiny{(0.27)} & 93.63 \tiny{(0.32)} & 84.43 \tiny{(1.38)} & 72.18 \tiny{(0.43)} & 36.33 \tiny{(0.70)} & 76.84 \tiny{(0.58)} & 61.37 \tiny{(1.83)} \\
    \hline
\textbf{Qwen}&&&&&&&\\

    \hspace{0.2cm}Echo & 63.80 \tiny{(0.66)} & 40.28 \tiny{(1.62)} & 85.19 \tiny{(0.35)} & 65.80 \tiny{(0.97)} & 64.57 \tiny{(0.35)} & 28.49 \tiny{(0.68)} & 62.19 \tiny{(0.64)} & 42.57 \tiny{(1.63)} \\
    \hspace{0.2cm}SPILL (Qwen)&  \underline{73.66} \tiny{(0.66)} & \underline{53.44} \tiny{(1.57)} & \underline{90.62} \tiny{(0.22)} & \underline{75.73} \tiny{(0.99)} & \underline{68.15} \tiny{(0.27)} & \underline{31.45} \tiny{(0.74)} & \underline{69.06} \tiny{(0.34)} & \underline{49.09} \tiny{(1.68)} \\
    \\
    \hspace{0.2cm}Summarizer & 64.80 \tiny{(0.35)} & 42.92 \tiny{(1.41)} & 91.55 \tiny{(0.18)} & 77.54 \tiny{(0.94)} & 76.33 \tiny{(0.48)} & 39.08 \tiny{(0.90)} & 76.43 \tiny{(0.89)} & 61.91 \tiny{(2.35)} \\
    \hspace{0.2cm}SPILL (Qwen)& \underline{70.98} \tiny{(0.22)} & \underline{49.94} \tiny{(0.67)} & \underline{94.10} \tiny{(0.19)} & \underline{85.02} \tiny{(1.04)} & \underline{77.41} \tiny{(0.24)} & \underline{42.45} \tiny{(1.42)} & \underline{78.12}\tiny{(0.40)} & \underline{\textbf{65.33}} \tiny{(1.27)} \\
    \hline

\textbf{Llama}&&&&&&&\\
    \hspace{0.2cm}Echo & 68.40 \tiny{(0.46)} & 46.20 \tiny{(0.46)} & 87.03 \tiny{(0.47)} & 70.60 \tiny{(0.79)} & 68.19 \tiny{(0.48)} & 31.49 \tiny{(1.24)} & 61.62 \tiny{(0.76)} & 42.24 \tiny{(1.45)} \\
    \hspace{0.2cm}SPILL (Llama)& \underline{73.44} \tiny{(0.35)} & \underline{53.50} \tiny{(0.60)} & \underline{90.49} \tiny{(0.29)} & \underline{78.89} \tiny{(0.47)} & \underline{70.53} \tiny{(0.27)} & \underline{33.55} \tiny{(1.19)} & \underline{67.42} \tiny{(0.64)} & \underline{47.57} \tiny{(1.82)} \\
   &&&&&&&\\
    \hspace{0.2cm}Summarizer & 67.47 \tiny{(0.21)} & 43.99 \tiny{(1.19)} & 92.49 \tiny{(0.31)} & 81.26 \tiny{(1.27)} & \underline{76.51} \tiny{(0.19)} & 40.10 \tiny{(0.86)} & 74.67 \tiny{(0.66)} & 59.23 \tiny{(1.62)} \\
    \hspace{0.2cm}SPILL (Llama)& \underline{70.31} \tiny{(0.20)} & \underline{48.62} \tiny{(0.99)} & \underline{93.59} \tiny{(0.17)} & \underline{86.12} \tiny{(0.96)} & 76.26 \tiny{(0.48)} & \underline{40.59} \tiny{(1.25)} & \underline{76.20} \tiny{(0.47)} & \underline{63.19} \tiny{(1.77)} \\

    \hline
\textbf{Gemma}&&&&&&&\\
    \hspace{0.2cm}Echo & 71.20 \tiny{(0.45)} & 50.32 \tiny{(1.88)} & 90.13 \tiny{(0.24)} & 73.36 \tiny{(0.65)} & 71.24 \tiny{(0.18)} & 32.82 \tiny{(0.80)} & 70.51 \tiny{(0.93)} & 50.13 \tiny{(0.89)} \\
    \hspace{0.2cm}SPILL (Gemma)& \underline{79.37} \tiny{(0.51)} & \underline{60.25} \tiny{(1.94)} & \underline{93.77} \tiny{(0.09)} & \underline{82.97} \tiny{(1.32)} & \underline{74.74} \tiny{(0.34)} & \underline{38.70} \tiny{(1.38)} & \underline{76.24} \tiny{(0.40)} & \underline{58.75} \tiny{(0.73)} \\
   &&&&&&&\\
    \hspace{0.2cm}Summarizer &69.74 \tiny{(0.28)} & 47.16 \tiny{(1.14)} & 94.10 \tiny{(0.24)} & 83.67 \tiny{(0.74)} & 78.90 \tiny{(0.32)} & \textbf{\underline{45.14}} \tiny{(1.87)} & 77.83 \tiny{(0.58)} & 63.48 \tiny{(2.22)} \\
    \hspace{0.2cm}SPILL (Gemma)& \underline{75.38} \tiny{(0.22)} & \underline{55.12} \tiny{(0.65)} & \textbf{\underline{95.49}} \tiny{(0.08)} & \textbf{\underline{88.25}} \tiny{(0.70)} & \textbf{\underline{79.01}} \tiny{(0.48)} & 43.77 \tiny{(1.04)} & \textbf{\underline{79.11}} \tiny{(0.57)} & \underline{64.33} \tiny{(2.54)} \\
    \hline
  \end{tabular}

  \caption{Results for the four benchmarks. Scores are averages over 5 runs, with standard deviations shown in parentheses. Model names in bold denote the embedding model, with names in parentheses indicating the LLM used. Plain refers to directly using the embedding for clustering. Boldface numbers highlight the highest values globally, while underlined values indicate the best within each embedder. Results for ClusterLLM, and KeyphraseClust are taken directly from previous papers.}
  \miniskip
  \label{tab:main result}
\end{table*}

\subsection{Main results}

Tabel \ref{tab:main result} compares our proposed method, SPILL, with all baselines with different encoder and decoder embedders. For encoder embedders, we found that SPILL generally performs best with \textbf{Gemma}, followed by \textbf{Qwen}, and then \textbf{LLama}.

Additionally, SPILL in almost all settings performs better than directly using the embeddings. In particular, our proposed method achieves results comparable to ClusterLLM \citep{zhang2023clusterllm}, which 
performs fine-tuning on the embedders. For decoder embedders, our method shows better performance than \textbf{Echo} and \textbf{Summarizer} on most datasets, and even better than encoder embedders on some datasets. This suggests that decoder embedders without contrastive loss optimization can outperform encoder embedders and show self-improvement with SPILL.

\begin{table*}[t]
  \centering
  \setlength{\tabcolsep}{3pt}
  \footnotesize
  \shrink
    \begin{tabular}{lcc|cc|cc|cc}
    \hline
    &\multicolumn{2}{c|}{Bank 77} & \multicolumn{2}{c|}{Clinc150}  & \multicolumn{2}{c|}{Mtop} & \multicolumn{2}{c}{Massive}\\
     \cline{1-3} \cline{4-5} \cline{6-7} \cline{8-9}  
    & NMI & Acc & NMI & Acc & NMI & Acc& NMI & Acc\\
   \hline
   \textbf{E5-large}&&&&&&&\\
    \hspace{0.2cm}Plain & 77.19 \tiny{(0.34)} & 59.60 \tiny{(1.42)} & 91.27 \tiny{(0.38)} & 75.92 \tiny{(0.91)} & 70.87 \tiny{(0.23)} & 34.21 \tiny{(0.57)} & 71.38 \tiny{(0.55)} & 53.85 \tiny{(1.28)} \\
    \hspace{0.4cm}w\textbackslash 1st stage & 79.79 \tiny{(0.24)} & 66.16 \tiny{(0.80)} & 91.49 \tiny{(0.31)} & 81.59 \tiny{(0.69)} & 70.62 \tiny{(0.53)} & 33.37 \tiny{(0.93)} & 71.32 \tiny{(0.26)} & 55.70 \tiny{(1.13)} \\
    \hspace{0.6cm}w\textbackslash 2nd stage (Gemma) & 83.56 \tiny{(0.47)} & 70.25 \tiny{(1.56)} & 92.93 \tiny{(0.08)} & 83.18 \tiny{(0.78)} & 71.77 \tiny{(0.35)} & 36.83 \tiny{(1.10)} & 75.40 \tiny{(0.51)} & 60.28 \tiny{(1.81)} \\
    \hspace{0.6cm}w\textbackslash 2nd stage (Llama) & 80.94 \tiny{(0.29)} & 66.84 \tiny{(0.97)} & 91.41 \tiny{(0.07)} & 80.09 \tiny{(1.31)} & 70.52 \tiny{(0.51)} & 35.04 \tiny{(1.06)} & 72.74 \tiny{(0.44)} & 58.06 \tiny{(1.52)} \\
    \hspace{0.6cm}w\textbackslash 2nd stage (Qwen) & 83.64 \tiny{(0.26)} & 70.31 \tiny{(0.75)} & 92.67 \tiny{(0.31)} & 82.58 \tiny{(0.87)} & 71.20 \tiny{(0.36)} & 36.48 \tiny{(0.73)} & 74.46 \tiny{(0.57)} & 59.38 \tiny{(1.20)} \\
   \hline
    \textbf{Gemma}&&&&&&&\\
    \hspace{0.2cm}Echo & 71.20 \tiny{(0.45)} & 50.32 \tiny{(1.88)} & 90.13 \tiny{(0.24)} & 73.36 \tiny{(0.65)} & 71.24 \tiny{(0.18)} & 32.82 \tiny{(0.80)} & 70.51 \tiny{(0.93)} & 50.13 \tiny{(0.89)} \\
    \hspace{0.4cm}w\textbackslash 1st stage & 72.09 \tiny{(0.39)} & 51.55 \tiny{(0.94)} & 90.75 \tiny{(0.17)} & 79.46 \tiny{(0.41)} & 71.70 \tiny{(0.42)} & 34.17 \tiny{(1.13)} & 70.45 \tiny{(0.33)} & 49.17 \tiny{(0.54)} \\
    \hspace{0.6cm}w\textbackslash 2nd stage & 79.37 \tiny{(0.51)} & 60.25 \tiny{(1.94)} & 93.77 \tiny{(0.09)} & 82.97 \tiny{(1.32)} & 74.74 \tiny{(0.34)} & 38.70 \tiny{(1.38)} & 76.24 \tiny{(0.40)} & 58.75 \tiny{(0.73)} \\
&&&&&&&\\

    \hspace{0.2cm}Summarizer & 69.74 \tiny{(0.28)} & 47.16 \tiny{(1.14)} & 94.10 \tiny{(0.24)} & 83.67 \tiny{(0.74)} & 78.90 \tiny{(0.32)} & 45.14 \tiny{(1.87)} & 77.83 \tiny{(0.58)} & 63.48 \tiny{(2.22)} \\
    \hspace{0.4cm}w\textbackslash 1st stage & 70.98 \tiny{(0.36)} & 49.84 \tiny{(0.65)} & 94.28 \tiny{(0.08)} & 86.64 \tiny{(0.41)} & 77.89 \tiny{(0.55)} & 41.65 \tiny{(1.31)} & 77.82 \tiny{(0.25)} & 63.38 \tiny{(1.14)} \\
    \hspace{0.6cm}w\textbackslash 2nd stage & 75.38 \tiny{(0.22)} & 55.12 \tiny{(0.65)} & 95.49 \tiny{(0.08)} & 88.25 \tiny{(0.70)} & 79.01 \tiny{(0.48)} & 43.77 \tiny{(1.04)} & 79.11 \tiny{(0.57)} & 64.33 \tiny{(2.54)} \\    

    \hline

  \end{tabular}
  

  \caption{Ablation study for the first and selection stage. Results are average over 5 runs. The results of other embedders are in the Appendix \ref{other_embedder}.}
  \miniskip
  \label{tab:contribution_1st2nd_stage}
\end{table*}

\subsection{Ablation and analysis}

\paragraph{Effectiveness of the first and second stage.} We analyze how the first and second stages improve clustering performance. Based on the results shown in Table \ref{tab:contribution_1st2nd_stage}, we find that the first selection stage generally improves performance compared to the plain setting, and the second selection stage often provides further performance gains compared to the first stage.

\paragraph{Analysis of Performance Variations.}

Although Table \ref{tab:main result} shows that our proposed method mostly improves performance than directly using an embedder, we analyze why some results still show lower performance. We found a higher correct selection ratio is associated with better clustering performance (see details in the Appendix \ref{other_embedder}).  We hypothesize that reduced performance occurs because some pooled utterances come from different clusters than the seed utterance. 

To verify our hypothesis, Table \ref{tab:oracle} shows a hypothetical result with $100\%$ correct selection rate. In this scenario, we assume that the correct candidates in the first stage are already known, without relying on an LLM for validation. Under this assumption, the results show consistent improvements over directly using the embedder, which supports our hypothesis, and aligns with the simulation result.



\begin{table*}[t]
  \centering
    \setlength{\tabcolsep}{2pt}

  \footnotesize
    \begin{tabular}{lcc|cc|cc|cc}
    \hline
    &\multicolumn{2}{c|}{Bank 77} & \multicolumn{2}{c|}{Clinc150}  & \multicolumn{2}{c|}{Mtop} & \multicolumn{2}{c}{Massive}\\
     \cline{1-3} \cline{4-5} \cline{6-7} \cline{8-9}  
    & NMI & Acc & NMI & Acc & NMI & Acc& NMI & Acc\\
    \hline
    \textbf{E5-large}&&&&&&&\\
    \hspace{0.2cm}Plain & 77.19 \tiny{(0.34)} & 59.60 \tiny{(1.42)} & 91.27 \tiny{(0.38)} & 75.92 \tiny{(0.91)} & 70.87 \tiny{(0.23)} & 34.21 \tiny{(0.57)} & 71.38 \tiny{(0.55)} & 53.85 \tiny{(1.28)} \\
    
    \hspace{0.2cm}Ground truth & 91.74 \tiny{(0.15)} & 83.40 \tiny{(0.97)} & 98.17 \tiny{(0.17)} & 94.28 \tiny{(1.11)} & 81.87 \tiny{(0.50)} & 45.44 \tiny{(1.65)} & 87.08 \tiny{(0.39)} & 73.16 \tiny{(1.87)} \\
    \hline

    \textbf{Gemma}&&&&&&&\\
    \hspace{0.2cm}Echo & 71.20 \tiny{(0.45)} & 50.32 \tiny{(1.88)} & 90.13 \tiny{(0.24)} & 73.36 \tiny{(0.65)} & 71.24 \tiny{(0.18)} & 32.82 \tiny{(0.80)} & 70.51 \tiny{(0.93)} & 50.13 \tiny{(0.89)} \\
    \hspace{0.2cm}Ground truth & 86.47 \tiny{(0.46)} & 71.49 \tiny{(0.88)} & 97.76 \tiny{(0.19)} & 92.21 \tiny{(0.85)} & 82.06 \tiny{(0.62)} & 45.74 \tiny{(1.85)} & 86.37 \tiny{(0.54)} & 69.00 \tiny{(1.76)} \\
    &&&&&&&\\
    \hspace{0.2cm}Sum & 69.74 \tiny{(0.28)} & 47.16 \tiny{(1.14)} & 94.10 \tiny{(0.24)} & 83.67 \tiny{(0.74)} & 78.90 \tiny{(0.32)} & 45.14 \tiny{(1.87)} & 77.83 \tiny{(0.58)} & 63.48 \tiny{(2.22)} \\
    \hspace{0.2cm}Ground truth & 81.05 \tiny{(0.33)} & 62.23 \tiny{(1.59)} & 97.68 \tiny{(0.09)} & 92.67 \tiny{(0.46)} & 83.65 \tiny{(0.44)} & 49.84 \tiny{(1.17)} & 85.94 \tiny{(0.40)} & 73.34 \tiny{(1.14)} \\  
    \hline

  \end{tabular}

  \caption{Average results of ground truth pooling (\%) over 5 runs. Ground truth pools each seed utterance with same-cluster utterances from 20 candidates in stage one. Results for other embedders are in the Appendix \ref{other_embedder}.}
  \miniskip
  \label{tab:oracle}
\end{table*}

\subsection{Qualitative analysis}
\begin{figure*}[ht!]
    \centering
    \begin{subfigure}[b]{0.48\textwidth}
        \includegraphics[width=\textwidth]{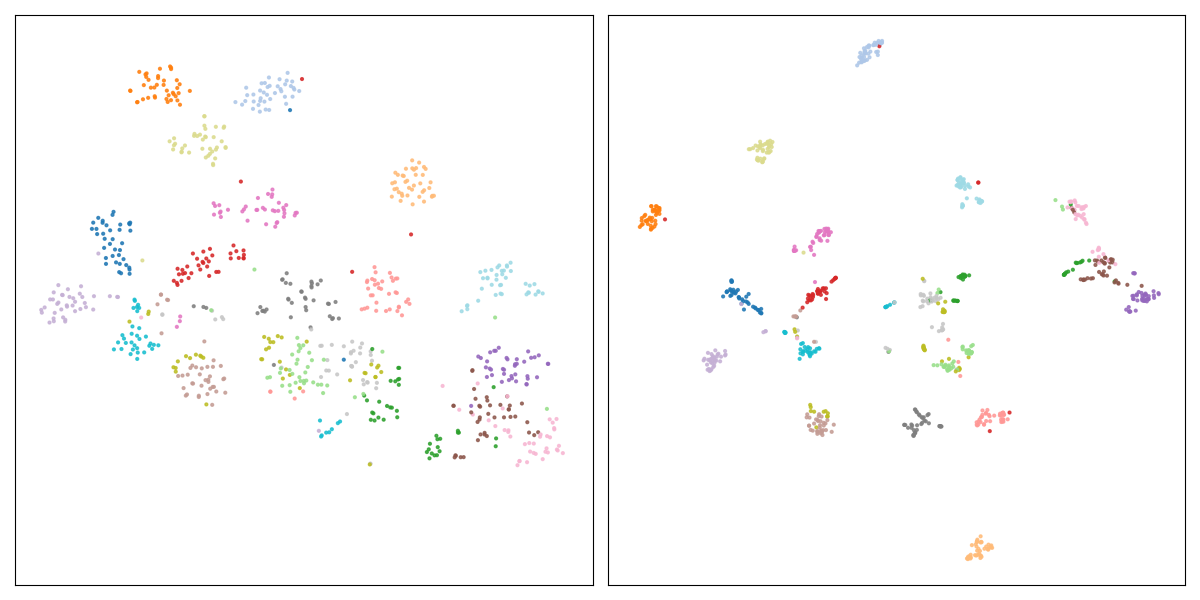}
        \caption{Bank77}
    \end{subfigure}
    \begin{subfigure}[b]{0.48\textwidth}
        \includegraphics[width=\textwidth]{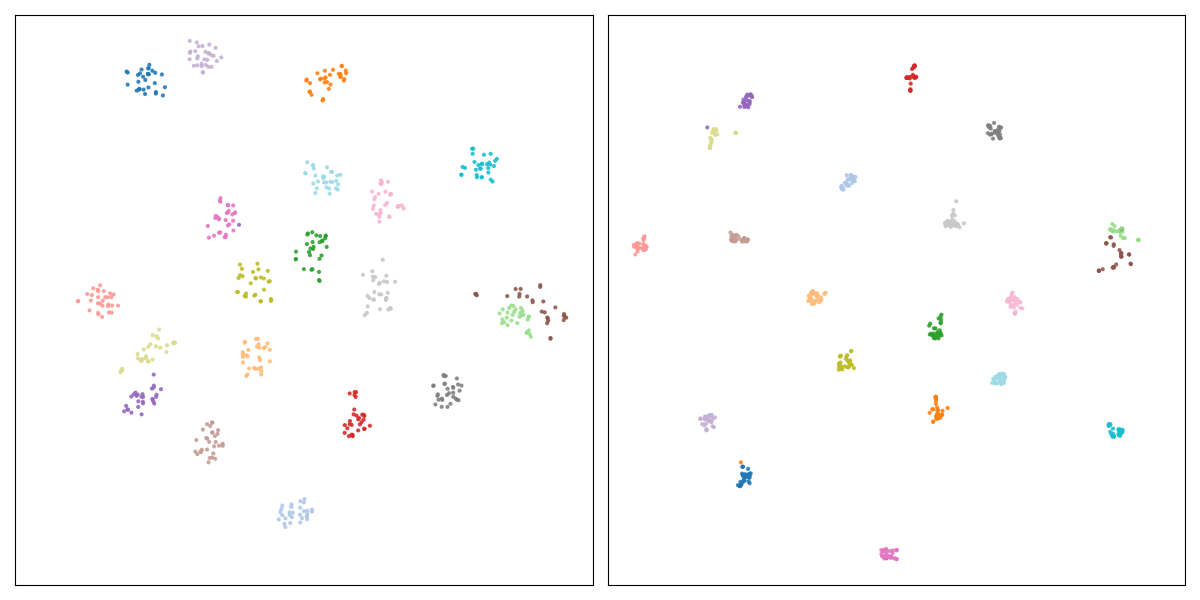}
        \caption{Clinc150}
    \end{subfigure}
    \begin{subfigure}[b]{0.48\textwidth}
        \includegraphics[width=\textwidth]{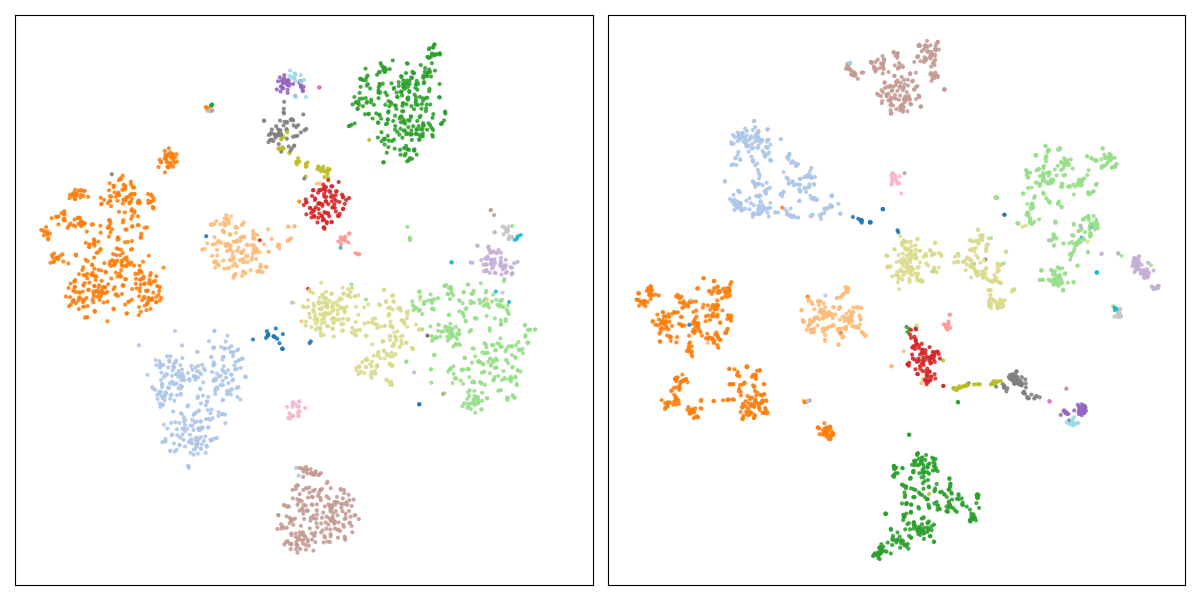}
        \caption{Mtop}
        \label{fig:graph3}
    \end{subfigure}
    \begin{subfigure}[b]{0.48\textwidth}
        \includegraphics[width=\textwidth]{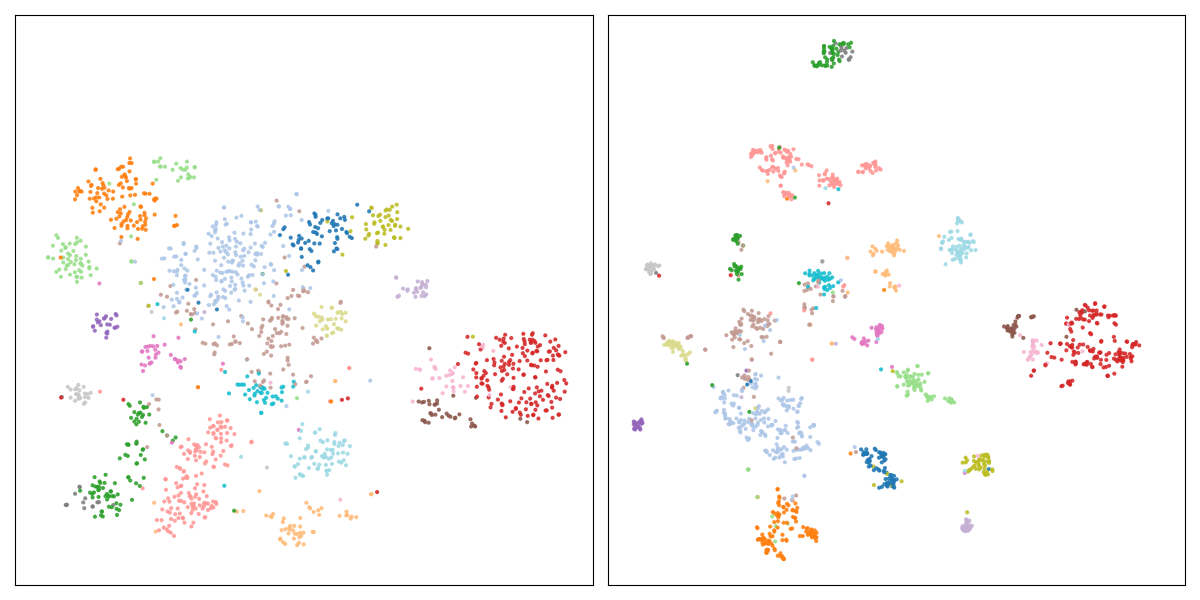}
        \caption{Massive}
        \label{fig:graph4}
    \end{subfigure}
    \shrink
    \caption{T-SNE plots for the four datasets (20 clusters for each). Left: Echo (Gemma). Right: SPILL}
    \label{fig:TSNE}
\end{figure*}

We apply t-SNE \citep{van2008visualizing} to reduce the embedding dimensions for 2D visualization. The embeddings are obtained from Gemma with Echo prompt. We compare the results between Echo and SPILL. From Figure~\ref{fig:TSNE}, it is clear that our method separates the clusters better. However, we also observe that for the Mtop dataset, both approaches struggle with clustering it well. As an additional analysis, we randomly select examples from the first and second stage selections using Gemma-Echo on the Bank77 dataset. In the first stage, we observe that utterances with different intents tend to appear as they become farther from the seed. Second, the LLM can identify utterances with the same intent as the seed. Finally, we find that the LLM is able to select more distant same-cluster utterances from the seed, thereby introducing greater variety.   The examples are provided in the Appendix \ref{qualatative_data}.

\section{Conclusions}
We proposed Selection and Pooling with Large Language Models (SPILL), an intuitive and domain-adaptive method for intent clustering without fine-tuning. SPILL is applicable to any embedder and does not require fine-tuning. We found that: (1) our proposed method enables viewing a clustering task as a small-scale selection problem, providing a novel perspective on the clustering process; (2) our method demonstrates general improvements regardless of the embedding model used, highlighting its versatility; (3) our method can achieve comparable results to other SOTA research studies without fine-tuning and by using smaller models; (4) it proves to be effective in data-limited domains, showing its adaptability in challenging scenarios. For future work, we see potential for improved prompting, the addition of few-shot settings, and generalisation to other languages. 

\section*{Acknowledgments}
This publication is part of the project LESSEN with project number NWA.1389.20.183 of the research program NWA ORC 2020/21 which is (partly) financed by the Dutch Research Council (NWO).


\section*{Limitations}
\textbf{Language.} Like most of the prior work, we only focus on English utterance datasets. This is relevant because most of benchmark datasets are in English.\\
\noindent
\textbf{Prompt design} Our research uses only one prompt in the second stage across different models and datasets. Our main goal is to avoid overdoing prompt engineering and to make our approach generalizable. The prompt simply explains the task and ensures the desired answer format for easy extraction. However, this also implies there still be room for improvement in our results because we did not specifically design a unique prompt for each LLM or dataset.\\
\noindent
\textbf{Few-shot settings.} Our research focuses solely on the scenario of an unlabeled dataset. However, in the second stage of selection. However, in our second stage selection, we use LLMs for selection. If LLMs are provided with some examples with known intents, they could potentially identify the relevance between seed and candidate utterances more effectively. This exploration will be left for future work.

\clearpage
\bibliographystyle{acl_natbib}
\bibliography{anthology,custom}
\clearpage

\appendix

\section*{Appendix}
\begin{figure*}[ht!]
    \centering
    \begin{subfigure}[b]{0.32\textwidth}
        \includegraphics[width=\textwidth]{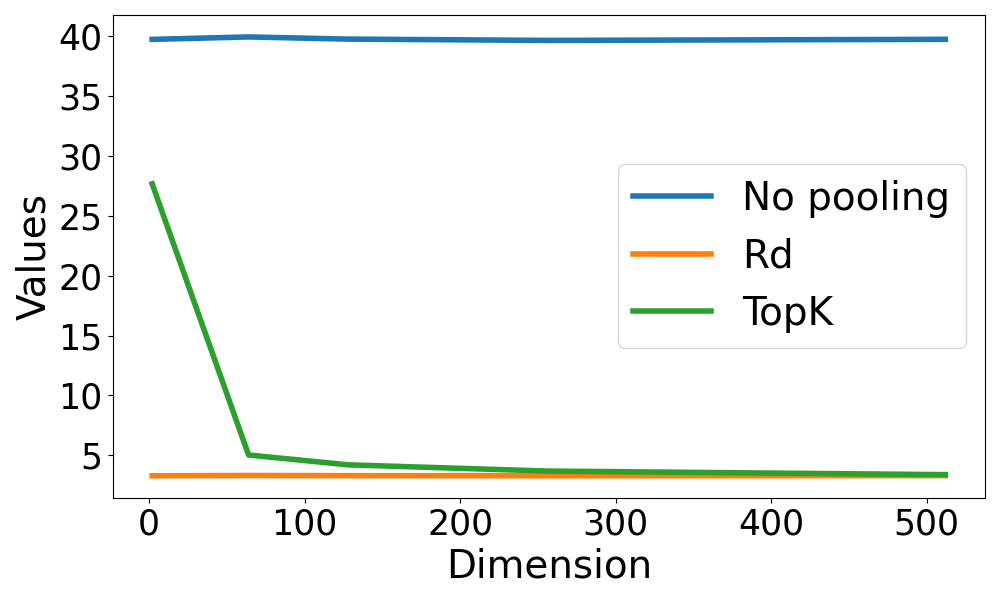}
        \caption{Estimated variance (normal)}
        \label{fig:graph1}
    \end{subfigure}
    \hfill
    \begin{subfigure}[b]{0.32\textwidth}
        \includegraphics[width=\textwidth]{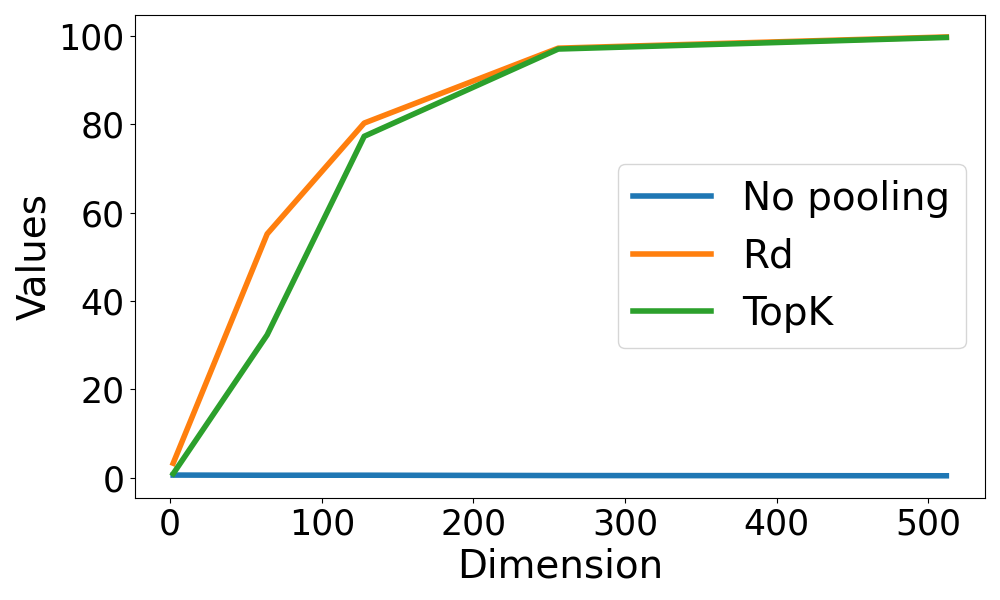}
        \caption{NMI (normal)}
        \label{fig:graph2}
    \end{subfigure}
    \hfill
    \begin{subfigure}[b]{0.32\textwidth}
        \includegraphics[width=\textwidth]{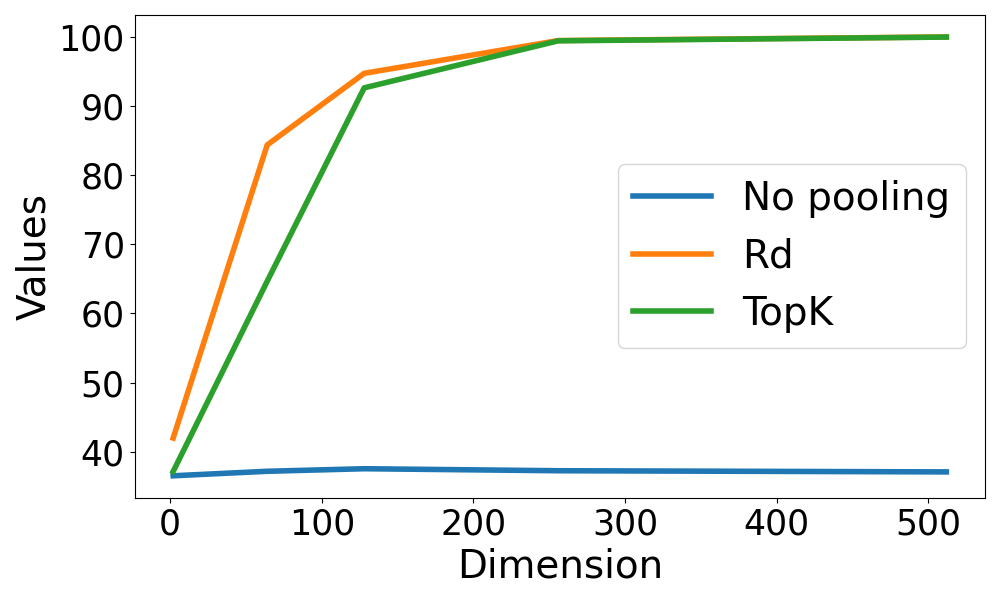}
        \caption{Acc (normal)}
    \end{subfigure}
    \vspace{0.5cm}
    \begin{subfigure}[b]{0.32\textwidth}
        \includegraphics[width=\textwidth]{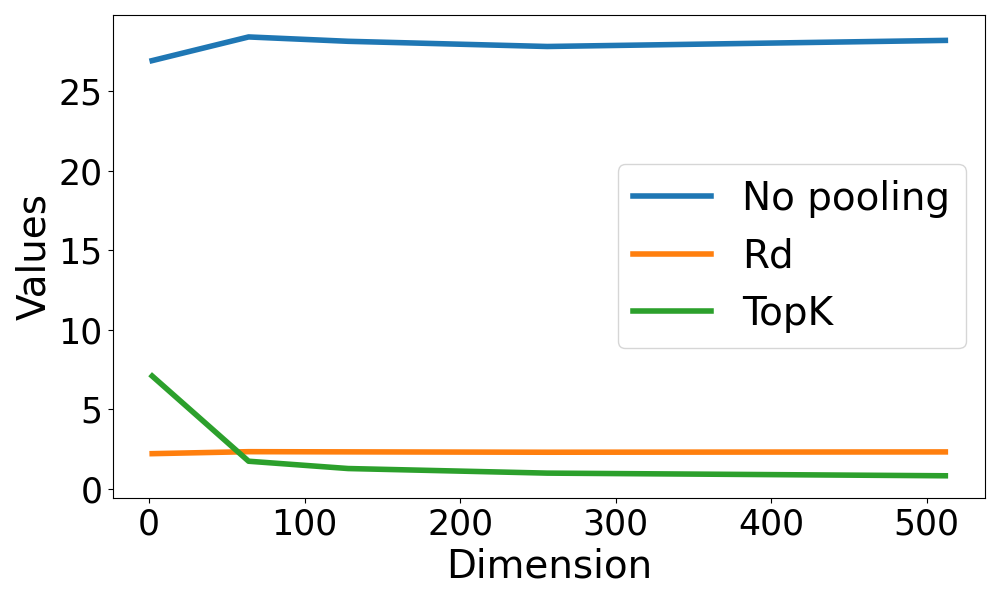}
        \caption{Estimated Variance (log-normal)}
    \end{subfigure}
    \hfill
    \begin{subfigure}[b]{0.32\textwidth}
        \includegraphics[width=\textwidth]{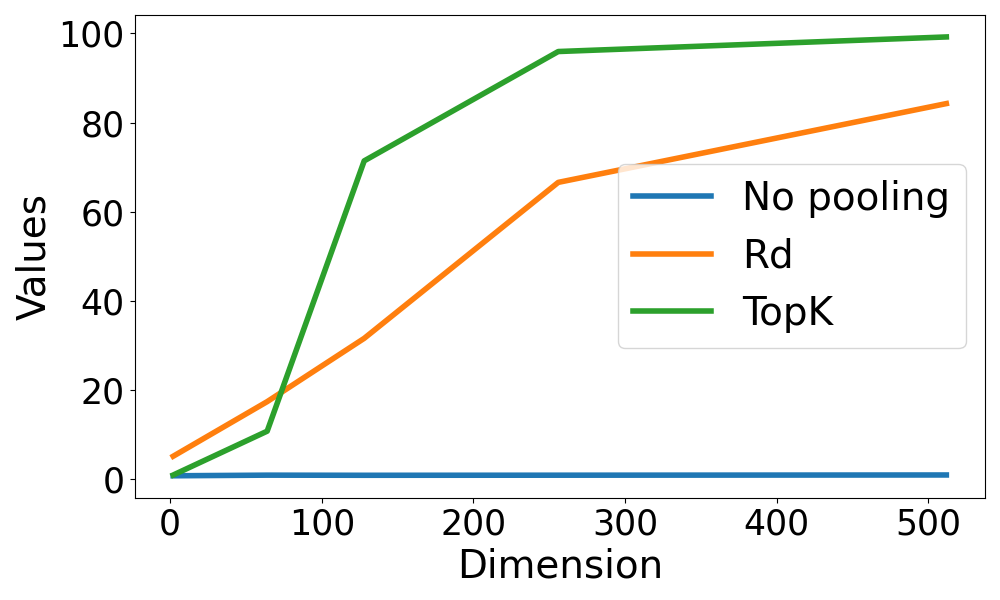}
        \caption{NMI (log-normal)}
    \end{subfigure}
    \hfill
    \begin{subfigure}[b]{0.32\textwidth}
        \includegraphics[width=\textwidth]{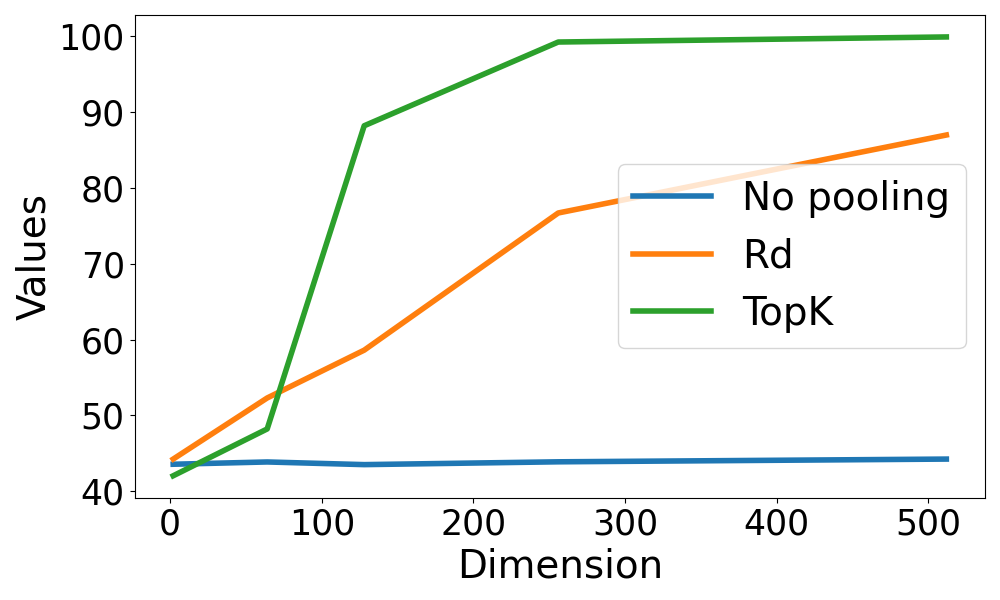}
        \caption{Acc (log-normal)}
    \end{subfigure}
    \caption{Average variance, clustering accuracy (Acc), and NMI over 50 runs with different dimensions given $k = 10$. We only present variance from one of the three clusters (they all have the same trend) for readability. The simulation process is the same as Table \ref{tab:simulation}, and only varies in dimension.}
    \label{fig:all_simu_plt}
\end{figure*}

\section{Simulation with varied dimensionality}\label{simu_dim}

Figure \ref{fig:all_simu_plt} shows \textbf{TopK} pooling performs even better in higher dimension than \textbf{Rd} when the cluster is skewed. For a normal distribution, \textbf{Rd} consistently outperforms \textbf{TopK}.

\section{Hyperparameter selection process}\label{hyperparameter}

We use \textbf{Instructor} as our first-stage embedding model to align with previous research, such as \citet{zhang2023clusterllm} and \citet{viswanathan-etal-2024-large}, who proposed methods built on this embedder. For the second stage, while both studies use \textbf{GPT}, we instead use the smaller and open-source model Llama for the reproducibility and accessibility.

Although a higher tolerance gives us a better chance to cover more utterances from the same cluster, this also increases the computation required for the second stage. We set $L = 20$.

\begin{figure}[t]
        \centering
        \includegraphics[width=0.5\textwidth]{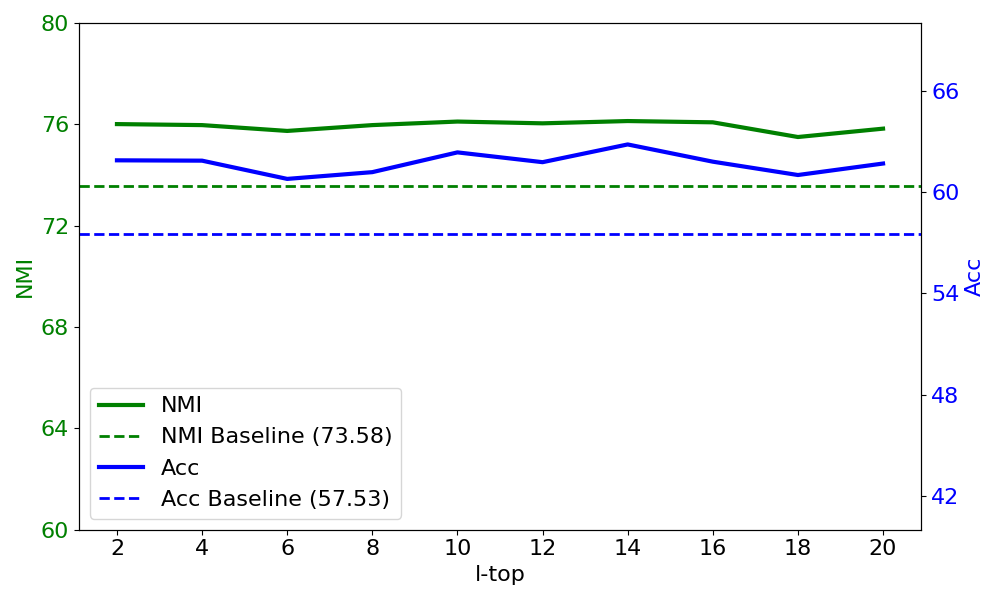}
        \caption{The average clustering accuracy (Acc) and NMI for different values of $l_{top}$ over 5 runs. The dashed line refers to the result obtained by directly using the embedder.}
        \label{fig:nmi_acc_sgd}
\end{figure}

We experiment with different combinations of $l_{top} \in \{2, 4, 6, 8, 10, 12, 14, 16, 18, 20\}$ utterances. Figure \ref{fig:nmi_acc_sgd} shows that the $(l_{top},l_{random}) = (14,6)$ overall gives the best performance. 

\section{Ablation analysis for all encoder and decoder embedders}\label{other_embedder}

Here, we provide the complete results of the ablation study. Details of the performance contributions in both the first and selection stages can be found in Table \ref{tab:contribution_1st2nd_stage_app}. The average correct ratio and the number of selected items are presented in Table  \ref{tab:correct_ratio_app}, while the details of the ground truth analysis can be found in Table \ref{tab:oracle_app}.

\begin{table*}[t]
  \centering
  \setlength{\tabcolsep}{3pt}
   \scriptsize

    \begin{tabular}{lcc|cc|cc|cc}
    \hline
    &\multicolumn{2}{c|}{Bank 77} & \multicolumn{2}{c|}{Clinc150}  & \multicolumn{2}{c|}{Mtop} & \multicolumn{2}{c}{Massive}\\
     \cline{1-3} \cline{4-5} \cline{6-7} \cline{8-9}  
    & NMI & Acc & NMI & Acc & NMI & Acc& NMI & Acc\\
    \hline
   \textbf{E5-large}&&&&&&&\\
    \hspace{0.2cm}Plain & 77.19 \tiny{(0.34)} & 59.60 \tiny{(1.42)} & 91.27 \tiny{(0.38)} & 75.92 \tiny{(0.91)} & 70.87 \tiny{(0.23)} & 34.21 \tiny{(0.57)} & 71.38 \tiny{(0.55)} & 53.85 \tiny{(1.28)} \\
    \hspace{0.4cm}w\textbackslash 1st stage & 79.79 \tiny{(0.24)} & 66.16 \tiny{(0.80)} & 91.49 \tiny{(0.31)} & 81.59 \tiny{(0.69)} & 70.62 \tiny{(0.53)} & 33.37 \tiny{(0.93)} & 71.32 \tiny{(0.26)} & 55.70 \tiny{(1.13)} \\
    \hspace{0.6cm}w\textbackslash 2nd stage (Gemma) & 83.56 \tiny{(0.47)} & 70.25 \tiny{(1.56)} & 92.93 \tiny{(0.08)} & 83.18 \tiny{(0.78)} & 71.77 \tiny{(0.35)} & 36.83 \tiny{(1.10)} & 75.40 \tiny{(0.51)} & 60.28 \tiny{(1.81)} \\
    \hspace{0.6cm}w\textbackslash 2nd stage (Llama) & 80.94 \tiny{(0.29)} & 66.84 \tiny{(0.97)} & 91.41 \tiny{(0.07)} & 80.09 \tiny{(1.31)} & 70.52 \tiny{(0.51)} & 35.04 \tiny{(1.06)} & 72.74 \tiny{(0.44)} & 58.06 \tiny{(1.52)} \\
    \hspace{0.6cm}w\textbackslash 2nd stage (Qwen) & 83.64 \tiny{(0.26)} & 70.31 \tiny{(0.75)} & 92.67 \tiny{(0.31)} & 82.58 \tiny{(0.87)} & 71.20 \tiny{(0.36)} & 36.48 \tiny{(0.73)} & 74.46 \tiny{(0.57)} & 59.38 \tiny{(1.20)} \\
   \hline
    \textbf{Instructor-large}&&&&&&&\\
    \hspace{0.2cm}Plain & 82.38 \tiny{(0.59)} & 65.70 \tiny{(1.78)} & 93.25 \tiny{(0.32)} & 81.12 \tiny{(2.27)} & 71.69 \tiny{(0.60)} & 34.06 \tiny{(2.51)} & 74.56 \tiny{(0.37)} & 56.62 \tiny{(1.75)} \\
    \hspace{0.4cm}w\textbackslash 1st stage & 82.83 \tiny{(0.45)} & 68.69 \tiny{(1.80)} & 93.48 \tiny{(0.17)} & 86.18 \tiny{(0.72)} & 71.65 \tiny{(0.34)} & 32.63 \tiny{(1.27)} & 75.29 \tiny{(0.59)} & 59.35 \tiny{(3.06)} \\
    \hspace{0.6cm}w\textbackslash 2nd stage (Gemma) & 85.01 \tiny{(0.29)} & 71.05 \tiny{(0.83)} & 93.77 \tiny{(0.35)} & 85.14 \tiny{(1.04)} & 72.65 \tiny{(0.32)} & 37.11 \tiny{(1.62)} & 77.62 \tiny{(0.46)} & 62.42 \tiny{(2.06)} \\
    \hspace{0.6cm}w\textbackslash 2nd stage (Llama) & 83.37 \tiny{(0.22)} & 69.55 \tiny{(0.60)} & 92.96 \tiny{(0.18)} & 84.31 \tiny{(0.72)} & 71.41 \tiny{(0.31)} & 35.18 \tiny{(1.12)} & 75.28 \tiny{(0.83)} & 58.79 \tiny{(2.75)} \\
    \hspace{0.6cm}w\textbackslash 2nd stage (Qwen) & 85.12 \tiny{(0.30)} & 71.48 \tiny{(0.27)} & 93.63 \tiny{(0.32)} & 84.43 \tiny{(1.38)} & 72.18 \tiny{(0.43)} & 36.33 \tiny{(0.70)} & 76.84 \tiny{(0.58)} & 61.37 \tiny{(1.83)} \\

    \hline

    \textbf{Qewn} &&&&&&\\
    \hspace{0.2cm}Echo & 63.80 \tiny{(0.66)} & 40.28 \tiny{(1.62)} & 85.19 \tiny{(0.35)} & 65.80 \tiny{(0.97)} & 64.57 \tiny{(0.35)} & 28.49 \tiny{(0.68)} & 62.19 \tiny{(0.64)} & 42.57 \tiny{(1.63)} \\
    \hspace{0.4cm}w\textbackslash 1st stage & 65.10 \tiny{(0.50)} & 42.84 \tiny{(0.85)} & 87.06 \tiny{(0.25)} & 70.95 \tiny{(1.05)} & 65.79 \tiny{(0.13)} & 29.43 \tiny{(0.26)} & 63.88 \tiny{(0.58)} & 43.69 \tiny{(1.73)} \\
    \hspace{0.6cm}w\textbackslash 2nd stage & 73.66 \tiny{(0.66)} & 53.44 \tiny{(1.57)} & 90.62 \tiny{(0.22)} & 75.73 \tiny{(0.99)} & 68.15 \tiny{(0.27)} & 31.45 \tiny{(0.74)} & 69.06 \tiny{(0.34)} & 49.09 \tiny{(1.68)} \\
    \\
    \hspace{0.2cm}Summarizer & 64.80 \tiny{(0.35)} & 42.92 \tiny{(1.41)} & 91.55 \tiny{(0.18)} & 77.54 \tiny{(0.94)} & 76.33 \tiny{(0.48)} & 39.08 \tiny{(0.90)} & 76.43 \tiny{(0.89)} & 61.91 \tiny{(2.35)} \\
    \hspace{0.4cm}w\textbackslash 1st stage & 65.21 \tiny{(0.39)} & 43.95 \tiny{(0.47)} & 92.35 \tiny{(0.15)} & 83.69 \tiny{(0.56)} & 76.13 \tiny{(0.25)} & 39.21 \tiny{(1.11)} & 76.66 \tiny{(0.49)} & 63.67 \tiny{(2.03)} \\
    \hspace{0.6cm}w\textbackslash 2nd stage & 70.98 \tiny{(0.22)} & 49.94 \tiny{(0.67)} & 94.10 \tiny{(0.19)} & 85.02 \tiny{(1.04)} & 77.41 \tiny{(0.24)} & 42.45 \tiny{(1.42)} & 78.12 \tiny{(0.40)} & 65.33 \tiny{(1.27)} \\
    \hline

    \textbf{Llama}&&&&&&&\\
    \hspace{0.2cm}Echo & 68.40 \tiny{(0.46)} & 46.20 \tiny{(0.46)} & 87.03 \tiny{(0.47)} & 70.60 \tiny{(0.79)} & 68.19 \tiny{(0.48)} & 31.49 \tiny{(1.24)} & 61.62 \tiny{(0.76)} & 42.24 \tiny{(1.45)} \\
    \hspace{0.4cm}w\textbackslash 1st stage & 70.25 \tiny{(0.56)} & 48.88 \tiny{(1.03)} & 87.57 \tiny{(0.14)} & 74.09 \tiny{(0.48)} & 68.91 \tiny{(0.42)} & 32.07 \tiny{(0.49)} & 63.19 \tiny{(0.40)} & 42.97 \tiny{(0.92)} \\
    \hspace{0.6cm}w\textbackslash 2nd stage & 73.44 \tiny{(0.35)} & 53.50 \tiny{(0.60)} & 90.49 \tiny{(0.29)} & 78.89 \tiny{(0.47)} & 70.53 \tiny{(0.27)} & 33.55 \tiny{(1.19)} & 67.42 \tiny{(0.64)} & 47.57 \tiny{(1.82)} \\
&&&&&&&\\

    \hspace{0.2cm}Summarizer & 67.47 \tiny{(0.21)} & 43.99 \tiny{(1.19)} & 92.49 \tiny{(0.31)} & 81.26 \tiny{(1.27)} & 76.51 \tiny{(0.19)} & 40.10 \tiny{(0.86)} & 74.67 \tiny{(0.66)} & 59.23 \tiny{(1.62)} \\
    \hspace{0.4cm}w\textbackslash 1st stage & 68.54 \tiny{(0.34)} & 46.11 \tiny{(0.77)} & 93.15 \tiny{(0.11)} & 85.55 \tiny{(0.23)} & 75.87 \tiny{(0.32)} & 39.41 \tiny{(1.34)} & 75.81 \tiny{(0.33)} & 63.30 \tiny{(1.82)} \\
    \hspace{0.6cm}w\textbackslash 2nd stage & 70.31 \tiny{(0.20)} & 48.62 \tiny{(0.99)} & 93.59 \tiny{(0.17)} & 86.12 \tiny{(0.96)} & 76.26 \tiny{(0.48)} & 40.59 \tiny{(1.25)} & 76.20 \tiny{(0.47)} & 63.19 \tiny{(1.77)} \\    
    \hline

    \textbf{Gemma}&&&&&&&\\
    \hspace{0.2cm}Echo & 71.20 \tiny{(0.45)} & 50.32 \tiny{(1.88)} & 90.13 \tiny{(0.24)} & 73.36 \tiny{(0.65)} & 71.24 \tiny{(0.18)} & 32.82 \tiny{(0.80)} & 70.51 \tiny{(0.93)} & 50.13 \tiny{(0.89)} \\
    \hspace{0.4cm}w\textbackslash 1st stage & 72.09 \tiny{(0.39)} & 51.55 \tiny{(0.94)} & 90.75 \tiny{(0.17)} & 79.46 \tiny{(0.41)} & 71.70 \tiny{(0.42)} & 34.17 \tiny{(1.13)} & 70.45 \tiny{(0.33)} & 49.17 \tiny{(0.54)} \\
    \hspace{0.6cm}w\textbackslash 2nd stage & 79.37 \tiny{(0.51)} & 60.25 \tiny{(1.94)} & 93.77 \tiny{(0.09)} & 82.97 \tiny{(1.32)} & 74.74 \tiny{(0.34)} & 38.70 \tiny{(1.38)} & 76.24 \tiny{(0.40)} & 58.75 \tiny{(0.73)} \\
&&&&&&&\\

    \hspace{0.2cm}Summarizer & 69.74 \tiny{(0.28)} & 47.16 \tiny{(1.14)} & 94.10 \tiny{(0.24)} & 83.67 \tiny{(0.74)} & 78.90 \tiny{(0.32)} & 45.14 \tiny{(1.87)} & 77.83 \tiny{(0.58)} & 63.48 \tiny{(2.22)} \\
    \hspace{0.4cm}w\textbackslash 1st stage & 70.98 \tiny{(0.36)} & 49.84 \tiny{(0.65)} & 94.28 \tiny{(0.08)} & 86.64 \tiny{(0.41)} & 77.89 \tiny{(0.55)} & 41.65 \tiny{(1.31)} & 77.82 \tiny{(0.25)} & 63.38 \tiny{(1.14)} \\
    \hspace{0.6cm}w\textbackslash 2nd stage & 75.38 \tiny{(0.22)} & 55.12 \tiny{(0.65)} & 95.49 \tiny{(0.08)} & 88.25 \tiny{(0.70)} & 79.01 \tiny{(0.48)} & 43.77 \tiny{(1.04)} & 79.11 \tiny{(0.57)} & 64.33 \tiny{(2.54)} \\

    \hline

  \end{tabular}
  
  \shrink

  \caption{Contribution of the first and selection stage (\%). Results averaged over 5 runs.}
  \miniskip
  \label{tab:contribution_1st2nd_stage_app}
\end{table*}

\begin{table*}[t]
  \centering
  \setlength{\tabcolsep}{2pt}
\scriptsize

    \begin{tabular}{lcc|cc|cc|cc}
    \hline
    &\multicolumn{2}{c|}{Bank 77} & \multicolumn{2}{c|}{Clinc150}  & \multicolumn{2}{c|}{Mtop} & \multicolumn{2}{c}{Massive}\\
     \cline{1-3} \cline{4-5} \cline{6-7} \cline{8-9}  
    & Ratio(\%)& $\#$ Selection & Ratio(\%)& $\#$ Selection & Ratio(\%)& $\#$ Selection & Ratio(\%)& $\#$ Selection  \\
    \hline
    \textbf{E5-large}&&&&&&&\\
    \hspace{0.2cm}Plain w\textbackslash 1st stage & 62.01 & 20.00 & 71.90 & 20.00 & 71.56 & 20.00 & 59.12 & 20.00 \\
    \hspace{0.4cm}w\textbackslash 2nd stage (Gemma) & 80.30 & 10.54 & 93.05 & 8.05 & 84.26 & 5.91 & 80.88 & 6.12 \\
    \hspace{0.4cm}w\textbackslash 2nd stage (Llama) & 71.14 & 12.13 & 86.52 & 9.50 & 80.12 & 7.09 & 73.42 & 7.38 \\
    \hspace{0.4cm}w\textbackslash 2nd stage (Qwen) & 80.52 & 8.61 & 90.80 & 7.14 & 82.45 & 4.72 & 78.65 & 5.17 \\
    \hline    
    
    \textbf{Instructor-large}&&&&&&&\\
    \hspace{0.2cm}w\textbackslash 1st stage & 67.84 & 20.00 & 78.17 & 20.00 & 75.41 & 20.00 & 65.38 & 20.00 \\
    \hspace{0.4cm}w\textbackslash 2nd stage (Gemma) & 80.66 & 11.28 & 93.43 & 8.15 & 85.46 & 5.94 & 81.85 & 6.27 \\
    \hspace{0.4cm}w\textbackslash 2nd stage (Llama) & 73.48 & 13.12 & 88.79 & 9.89 & 81.98 & 7.26 & 76.34 & 7.73 \\
    \hspace{0.4cm}w\textbackslash 2nd stage (Qwen) & 81.12 & 9.41 & 92.06 & 7.31 & 84.50 & 4.82 & 79.70 & 5.38 \\

    \hline
   \textbf{Qwen} &&&&&&&\\
    \hspace{0.2cm}Echo&&&&&&&\\  
    \hspace{0.4cm}w\textbackslash 1st stage & 44.11 & 20.00 & 65.69 & 20.00 & 70.18 & 20.00 & 56.66 & 20.00 \\
    \hspace{0.6cm}w\textbackslash 2st stage  & 72.99 & 6.94 & 90.22 & 6.24 & 83.29 & 3.84 & 78.96 & 4.25 \\
    
    \hspace{0.2cm}Summarizer&&&&&&&\\
    \hspace{0.4cm}w\textbackslash 1st stage& 43.44 & 20.00 & 78.49 & 20.00 & 83.28 & 20.00 & 71.47 & 20.00 \\
    \hspace{0.6cm}w\textbackslash 2st stage& 69.95 & 7.30 & 93.03 & 6.72 & 89.15 & 4.17 & 82.78 & 4.94 \\
    \hline

    \textbf{Llama}&&&&&&&\\
    \hspace{0.2cm}Echo&&&&&&&\\
    \hspace{0.4cm}w\textbackslash 1st stage & 48.89 & 20.00 & 66.84 & 20.00 & 73.51 & 20.00 & 55.20 & 20.00 \\
    \hspace{0.6cm}w\textbackslash 2nd stage & 61.90 & 10.54 & 85.84 & 8.41 & 82.83 & 6.23 & 71.70 & 6.32 \\
    \hspace{0.2cm}Summarizer&&&&&&&\\
    \hspace{0.4cm}w\textbackslash 1st stage & 47.85 & 20.00 & 81.76 & 20.00 & 83.46 & 20.00 & 71.66 & 20.00 \\
    \hspace{0.6cm}w\textbackslash 2nd stage & 57.70 & 11.54 & 89.90 & 9.58 & 87.09 & 6.97 & 79.37 & 7.42 \\

    \hline
    \textbf{Gemma}&&&&&&&\\
    \hspace{0.2cm}Echo&&&&&&&\\
    \hspace{0.4cm}w\textbackslash 1st stage & 50.85 & 20.00 & 71.15 & 20.00 & 77.01 & 20.00 & 61.25 & 20.00 \\
    \hspace{0.6cm} w\textbackslash 2nd stage & 75.12 & 9.60 & 93.92 & 7.54 & 88.46 & 4.94 & 82.02 & 5.45 \\
    \hspace{0.2cm}Summarizer&&&&&&&\\
    \hspace{0.4cm}w\textbackslash 1st stage & 52.11 & 20.00 & 84.68 & 20.00 & 85.34 & 20.00 & 74.49 & 20.00 \\
    \hspace{0.6cm}w\textbackslash 2nd stage & 73.00 & 10.10 & 95.32 & 8.01 & 90.19 & 5.43 & 84.60 & 6.13 \\ 

    \hline
  \end{tabular}
  
  \shrink

  \caption{Average correct selection ratio and average number of utterances selected over 5 runs. Note: the number for the first stage remains constant as $L = 20$ is fixed. The ratio is defined as the number of same-cluster utterances out of the total number of selections.}
  \miniskip
  \label{tab:correct_ratio_app}
\end{table*}

\begin{table*}[t]
  \centering
    \setlength{\tabcolsep}{2pt}

  \scriptsize
    \begin{tabular}{lcc|cc|cc|cc}
    \hline
    &\multicolumn{2}{c|}{Bank 77} & \multicolumn{2}{c|}{Clinc150}  & \multicolumn{2}{c|}{Mtop} & \multicolumn{2}{c}{Massive}\\
     \cline{1-3} \cline{4-5} \cline{6-7} \cline{8-9}  
    & NMI & Acc & NMI & Acc & NMI & Acc& NMI & Acc\\
    \hline
    \textbf{E5-large}&&&&&&&\\
    \hspace{0.2cm}Plain & 77.19 \tiny{(0.34)} & 59.60 \tiny{(1.42)} & 91.27 \tiny{(0.38)} & 75.92 \tiny{(0.91)} & 70.87 \tiny{(0.23)} & 34.21 \tiny{(0.57)} & 71.38 \tiny{(0.55)} & 53.85 \tiny{(1.28)} \\
    
    \hspace{0.2cm}Ground truth & 91.74 \tiny{(0.15)} & 83.40 \tiny{(0.97)} & 98.17 \tiny{(0.17)} & 94.28 \tiny{(1.11)} & 81.87 \tiny{(0.50)} & 45.44 \tiny{(1.65)} & 87.08 \tiny{(0.39)} & 73.16 \tiny{(1.87)} \\
    \hline
    \textbf{Instructor-large}&&&&&&&\\
    \hspace{0.2cm}Plain & 82.38 \tiny{(0.59)} & 65.70 \tiny{(1.78)} & 93.25 \tiny{(0.32)} & 81.12 \tiny{(2.27)} & 71.69 \tiny{(0.60)} & 34.06 \tiny{(2.51)} & 74.56 \tiny{(0.37)} & 56.62 \tiny{(1.75)} \\
    \hspace{0.2cm}Ground truth & 92.68 \tiny{(0.32)} & 84.54 \tiny{(0.91)} & 98.54 \tiny{(0.10)} & 95.44 \tiny{(0.40)} & 81.29 \tiny{(0.27)} & 43.69 \tiny{(1.10)} & 87.30 \tiny{(0.35)} & 72.04 \tiny{(1.87)} \\

    \hline
    \textbf{Qwen}&&&&&&&\\
       \hspace{0.2cm}Echo & 63.80 \tiny{(0.66)} & 40.28 \tiny{(1.62)} & 85.19 \tiny{(0.35)} & 65.80 \tiny{(0.97)} & 64.57 \tiny{(0.35)} & 28.49 \tiny{(0.68)} & 62.19 \tiny{(0.64)} & 42.57 \tiny{(1.63)} \\
    \hspace{0.2cm}Ground truth &  81.64 \tiny{(0.68)} & 64.21 \tiny{(1.87)} & 95.60 \tiny{(0.15)} & 87.15 \tiny{(0.68)} & 76.14 \tiny{(0.42)} & 38.35 \tiny{(1.25)} & 79.49 \tiny{(0.46)} & 60.56 \tiny{(1.22)} \\

      &&&&&&&\\

     \hspace{0.2cm}Sum & 64.80 \tiny{(0.35)} & 42.92 \tiny{(1.41)} & 91.55 \tiny{(0.18)} & 77.54 \tiny{(0.94)} & 76.33 \tiny{(0.48)} & 39.08 \tiny{(0.90)} & 76.43 \tiny{(0.89)} & 61.91 \tiny{(2.35)} \\
    \hspace{0.2cm}Ground truth & 77.05 \tiny{(0.56)} & 57.58 \tiny{(0.79)} & 97.33 \tiny{(0.17)} & 91.91 \tiny{(0.71)} & 83.19 \tiny{(0.40)} & 49.35 \tiny{(1.15)} & 86.99 \tiny{(0.33)} & 74.72 \tiny{(0.90)} \\
        \hline

    \textbf{Llama}&&&&&&&\\
    \hspace{0.2cm}Echo & 68.40 \tiny{(0.46)} & 46.20 \tiny{(0.46)} & 87.03 \tiny{(0.47)} & 70.60 \tiny{(0.79)} & 68.19 \tiny{(0.48)} & 31.49 \tiny{(1.24)} & 61.62 \tiny{(0.76)} & 42.24 \tiny{(1.45)} \\
    \hspace{0.2cm}Ground truth & 85.44 \tiny{(0.39)} & 68.84 \tiny{(0.82)} & 95.74 \tiny{(0.11)} & 88.45 \tiny{(0.75)} & 79.27 \tiny{(0.28)} & 40.69 \tiny{(0.79)} & 78.74 \tiny{(0.79)} & 59.48 \tiny{(2.75)} \\
    &&&&&&&\\
    \hspace{0.2cm}Sum & 67.47 \tiny{(0.21)} & 43.99 \tiny{(1.19)} & 92.49 \tiny{(0.31)} & 81.26 \tiny{(1.27)} & 76.51 \tiny{(0.19)} & 40.10 \tiny{(0.86)} & 74.67 \tiny{(0.66)} & 59.23 \tiny{(1.62)} \\
    \hspace{0.2cm}Ground truth & 80.08 \tiny{(0.30)} & 59.68 \tiny{(0.74)} & 97.56 \tiny{(0.08)} & 93.41 \tiny{(0.62)} & 82.77 \tiny{(0.29)} & 48.53 \tiny{(0.79)} & 85.66 \tiny{(0.64)} & 72.53 \tiny{(2.05)} \\ 
    \hline

    \textbf{Gemma}&&&&&&&\\
    \hspace{0.2cm}Echo & 71.20 \tiny{(0.45)} & 50.32 \tiny{(1.88)} & 90.13 \tiny{(0.24)} & 73.36 \tiny{(0.65)} & 71.24 \tiny{(0.18)} & 32.82 \tiny{(0.80)} & 70.51 \tiny{(0.93)} & 50.13 \tiny{(0.89)} \\
    \hspace{0.2cm}Ground truth & 86.47 \tiny{(0.46)} & 71.49 \tiny{(0.88)} & 97.76 \tiny{(0.19)} & 92.21 \tiny{(0.85)} & 82.06 \tiny{(0.62)} & 45.74 \tiny{(1.85)} & 86.37 \tiny{(0.54)} & 69.00 \tiny{(1.76)} \\
    &&&&&&&\\
    \hspace{0.2cm}Sum & 69.74 \tiny{(0.28)} & 47.16 \tiny{(1.14)} & 94.10 \tiny{(0.24)} & 83.67 \tiny{(0.74)} & 78.90 \tiny{(0.32)} & 45.14 \tiny{(1.87)} & 77.83 \tiny{(0.58)} & 63.48 \tiny{(2.22)} \\
    \hspace{0.2cm}Ground truth & 81.05 \tiny{(0.33)} & 62.23 \tiny{(1.59)} & 97.68 \tiny{(0.09)} & 92.67 \tiny{(0.46)} & 83.65 \tiny{(0.44)} & 49.84 \tiny{(1.17)} & 85.94 \tiny{(0.40)} & 73.34 \tiny{(1.14)} \\  
    \hline

  \end{tabular}
  \shrink

  \caption{Average results of ground truth pooling (\%) over 5 runs. Ground truth refers to each seed utterance being pooled with other same-cluster (i.e., always selected correctly from a pool of 20 candidates) utterances from the first-stage collection.}
  \miniskip
  \label{tab:oracle_app}
\end{table*}

\section{Some examples of Selection} \label{qualatative_data}
We randomly select three examples extracted from the first selection using \textbf{Gemma-Echo} and our method \textbf{Gemma-SPILL} from the \textbf{Bank77} dataset. In the first stage, the selection is ordered by the distance from the seed (starting at 15, derived through chunk sampling). First, we observe that in the initial stage, utterances with different intents appear in the later part of the selection. Second, it can be seen that the LLM effectively selects utterances with the same intent as the seed. Additionally, the LLM is capable of correctly selecting utterances from the later part of the list, introducing more variety to the pool while maintaining consistency with the seed. For instance, in Example 1, the furthest utterance is selected by the LLM without any mistakes in selecting incorrect utterances.
\\

\noindent \textbf{Example 1:}\\
Seed utterance: I would appreciate it if I could get an item refunded (intent: request refund)\\

\textbf{Gemma-Echo-1stStage}\\
\noindent1. Would I be able to get a refund for something I bought? (request refund)

\noindent2. I don't want the item, I bought it on accident, can I get a refund? (request refund)

\noindent3. I would like a refund for something I bought (request refund)

\noindent4. I want to get an item refunded (request refund)

\noindent5. I bought something but now I would like a refund. How do I do that? (request refund)

\noindent6. I am not Happy with this product can i get a refund? (request refund)

\noindent7. Can I have an item refunded? (request refund)

\noindent8. I bought this item and was charged the wrong amount can I get a refund? (request refund)

\noindent9. I am unhappy with my purchase, how do I cancel the order? (request refund)

\noindent10. I would like to cancel a payment. I purchased something several days ago and i still have not received it. (request refund)

\noindent11. I would like to know why I was charged twice for my purchase. (transaction charged twice)

\noindent12. I would like to know why my payment is still pending, can you help? (pending transfer)

\noindent13. Hello! I recently made a purchase and I'm needing to cancel my order and process a refund as soon as possible. (request refund)

\noindent14. I would like a refund on one of your products that has been sold to me (request refund)

\noindent15. Can I receive a refund for my item? (request refund)

\noindent16. Can I have a refund? (request refund)

\noindent17. I want to reverse a purchase. Can I cancel it? (request refund)

\noindent18. I need a refund on an item I have not received.  Am I able to simply cancel the payment?  I don't know how to do this. (request refund)

\noindent19. I requested a refund, and never received it. What can I do? (Refund not showing up)

\noindent20. Hi there! I need to cancel an order I recently made and start processing a refund. Can you please help me with this and set up the refund as soon as possible? It's very urgent. (request refund)
\\

\textbf{Gemma-Echo-2ndStage}\\
\noindent1. Can I have a refund? (request refund)

\noindent2. Can I receive a refund for my item? (request refund)

\noindent3. Would I be able to get a refund for something I bought? (request refund)

\noindent4. I would like a refund on one of your products that has been sold to me (request refund)

\noindent5. I am not Happy with this product can i get a refund? (request refund)

\noindent6. I need a refund on an item I have not received.  Am I able to simply cancel the payment?  I don't know how to do this. (request refund)

\noindent7. I want to get an item refunded (request refund)

\noindent8. I would like a refund for something I bought (request refund)

\noindent9. Hi there! I need to cancel an order I recently made and start processing a refund. Can you please help me with this and set up the refund as soon as possible? It's very urgent. (request refund)

\noindent10. I bought this item and was charged the wrong amount can I get a refund? (request refund)

\noindent11. I don't want the item, I bought it on accident, can I get a refund? (request refund)

\noindent12. Can I have an item refunded? (request refund)

\noindent13. I bought something but now I would like a refund. How do I do that? (request refund)
\\
\\
\noindent \textbf{Example 2:}\\
Seed utterance: When getting my ID checked, what are the steps involved? (intent: verify my identity)\\

\textbf{Gemma-Echo-1stStage}\\
\noindent1. What kind of documents do I need for the identity check? (verify my identity)

\noindent2. What will I need for identity verification? (verify my identity)

\noindent3. Are there any documents needed for the identity check? (verify my identity)

\noindent4. What's the process for ID verification? (verify my identity)

\noindent5. Is there any documentation needed for the identity check? (verify my identity)

\noindent6. What are the steps I need to take to verify my identity? (verify my identity)

\noindent7. Do I need any kind of documentation for the identity check? (verify my identity)

\noindent8. What do I do for the identity check? (verify my identity)

\noindent9. What is needed to prove my identity? (verify my identity)

\noindent10. Let me know the steps for the identity checks (verify my identity)

\noindent11. What is the need to verify my identity? (why verify identity)

\noindent12. What are the steps to verify my identity? (verify my identity)

\noindent13. What all am I required to show for the identity check? (verify my identity)

\noindent14. What things do I need to verify my identity? (verify my identity)

\noindent15. Is there a specific type you need for identity verification? (verify my identity)

\noindent16. What do I need to do to verify the source of my funds? (verify source of funds)

\noindent17. What's with not verifying my Id? (unable to verify identity)

\noindent18. I need the source of my funds verified. How do I do this? (verify source of funds)

\noindent19. What do i need to verify my id? (unable to verify identity)

\noindent20. What other methods are there to verify my identity? (why verify identity)
\\

\textbf{Gemma-Echo-2ndStage}\\

\noindent1. What are the steps I need to take to verify my identity? (verify my identity)

\noindent2. What's the process for ID verification? (verify my identity)

\noindent3. Let me know the steps for the identity checks (verify my identity)

\noindent4. What are the steps to verify my identity? (verify my identity)
\\

\end{document}